\documentclass[10pt,twocolumn,twoside]{IEEEtran}
\makeatletter
\long\def\@makecaption#1#2{\ifx\@captype\@IEEEtablestring%
\footnotesize\begin{center}{\normalfont\footnotesize #1}\\
{\normalfont\footnotesize\scshape #2}\end{center}%
\@IEEEtablecaptionsepspace
\else
\@IEEEfigurecaptionsepspace
\setbox\@tempboxa\hbox{\normalfont\footnotesize {#1.}~~ #2}%
\ifdim \wd\@tempboxa >\hsize%
\setbox\@tempboxa\hbox{\normalfont\footnotesize {#1.}~~ }%
\parbox[t]{\hsize}{\normalfont\footnotesize \noindent\unhbox\@tempboxa#2}%
\else
\hbox to\hsize{\normalfont\footnotesize\hfil\box\@tempboxa\hfil}\fi\fi}
\makeatother

\usepackage{cite}

\ifCLASSINFOpdf
\usepackage{multirow}
\usepackage{cite}
\usepackage{graphicx}
\usepackage{subfigure}
\usepackage{tabularx}
\usepackage{diagbox}
\usepackage{eqnarray}
\usepackage{multirow}
\usepackage{amsmath}
\usepackage{times}
\usepackage{helvet}
\usepackage{courier}
\usepackage{balance} 
\usepackage{verbatim}
\usepackage{amsfonts}
\usepackage{algorithm}
\usepackage{algorithmic} 
\usepackage{color}
\else
\fi
\begin{document}

%
\title{Object-Part Attention Driven Discriminative Localization for Fine-grained Image Classification}

\title{Object-Part Attention Model for \\Fine-grained Image Classification}



\author{Yuxin Peng, Xiangteng He, and Junjie Zhao
\thanks{This work was supported by National Natural Science Foundation of China under Grants 61771025, 61371128 and 61532005.}
\thanks{The authors are with the Institute of Computer Science and Technology, Peking University, Beijing 100871, China. Corresponding author: Yuxin Peng (e-mail: pengyuxin@pku.edu.cn).}}


%



\maketitle

\begin{abstract}
Fine-grained image classification is to recognize hundreds of subcategories belonging to the same basic-level category, such as 200 subcategories belonging to the bird, which is highly challenging due to \emph{large} variance in the same subcategory and \emph{small} variance among different subcategories.
Existing methods generally first locate the objects or parts and then discriminate which subcategory the image belongs to. 
However, they mainly have \emph{two limitations}: 
(1) Relying on object or part annotations which are heavily labor consuming. 
(2) Ignoring the spatial relationships between the object and its parts as well as among these parts, both of which are significantly helpful for finding discriminative parts.
Therefore, this paper proposes the \emph{object-part attention model (OPAM)} for weakly supervised fine-grained image classification, and the main novelties are:
(1) \emph{Object-part attention model} integrates two level attentions: 
\emph{object-level attention} localizes objects of images, and \emph{part-level attention} selects discriminative parts of object. 
Both are jointly employed to learn multi-view and multi-scale features to enhance their mutual promotions. 
(2) \emph{Object-part spatial constraint model} combines two spatial constraints: \emph{object spatial constraint} ensures selected parts highly representative, and \emph{part spatial constraint} eliminates redundancy and enhances discrimination of selected parts. 
Both are jointly employed to exploit the subtle and local differences for distinguishing the subcategories.  
Importantly, neither object nor part annotations are used in our proposed approach, which avoids the heavy labor consumption of labeling.   
Comparing with more than 10 state-of-the-art methods on 4 widely-used datasets, our OPAM approach achieves the best performance. 
\end{abstract}


%
\IEEEpeerreviewmaketitle

\section{Introduction}
%
%
%
%

\begin{figure}[!ht]
  \begin{center}\includegraphics[width=1\linewidth]{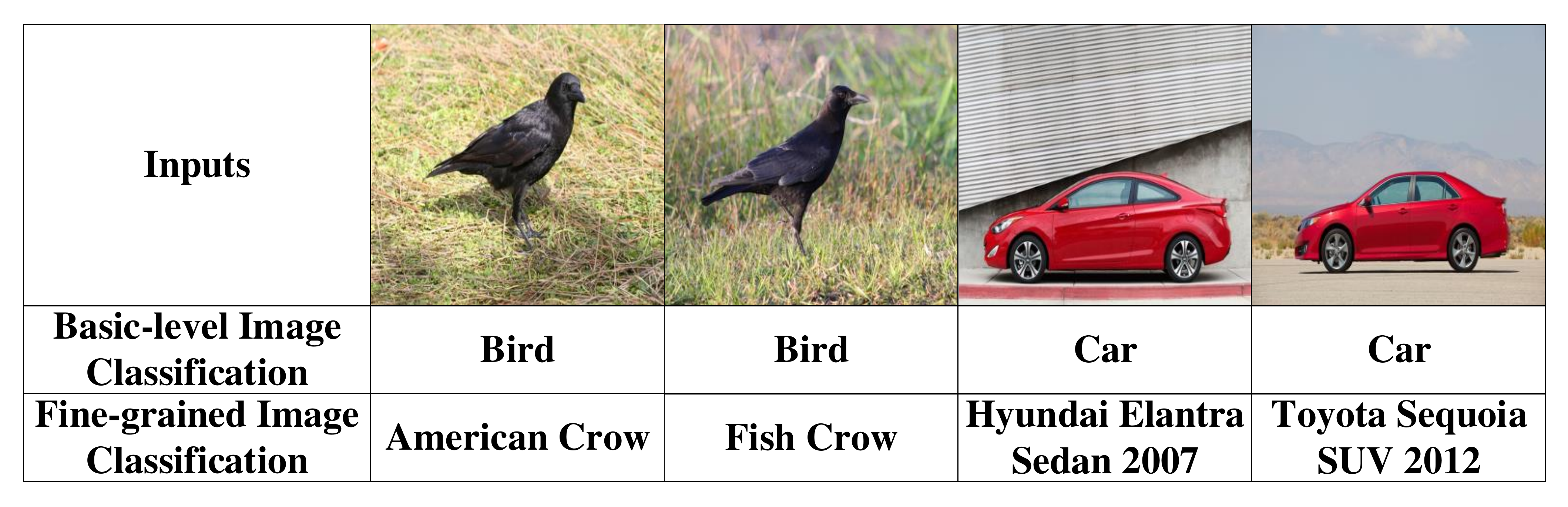}
  \caption{Basic-level image classification vs. fine-grained image classification. In basic-level image classification, we only need to classify the first two images as bird category, distinguishing them from car category. While in fine-grained image classification, the subcategory should be further determined exactly. For example, the first two images belong to the subcategories of American Crow and Fish Crow respectively.}
  \label{fine-grained}
  \end{center}
\end{figure}

\begin{figure*}[!ht]
  \begin{center}
    \includegraphics[width=1\linewidth]{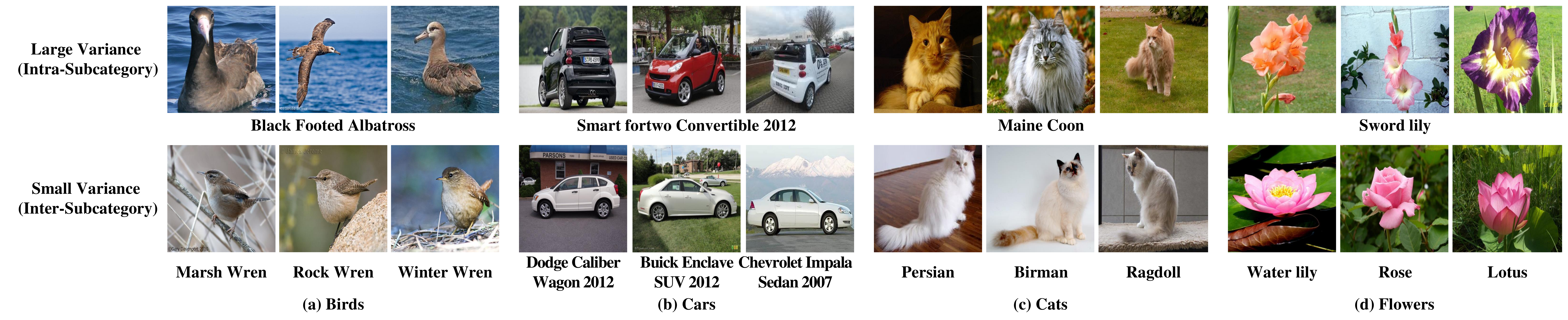}
    \caption{Illustration of challenges in fine-grained image classification: large variance in same subcategory as shown in the first row, and small variance among different subcategories as shown in the second row. The images in (a) Birds, (b) Cars, (c) Cats and (d) Flowers are from CUB-200-2011 \cite{wah2011caltech}, Cars-196 \cite{krause20133d}, Oxford-IIIT Pet \cite{parkhi2012cats} and Oxford-Flower-102 \cite{Nilsback2008Automated} datasets respectively.}
    \label{interintra}
  \end{center}
\end{figure*}

\IEEEPARstart{F}{ine-grained} image classification is highly challenging, aiming to recognize hundreds of subcategories under the same basic-level category, such as hundreds of subcategories of birds \cite{wah2011caltech}, cars \cite{krause20133d}, pets \cite{parkhi2012cats}, flowers \cite{Nilsback2008Automated} and aircrafts \cite{maji2013fine}. 
While basic-level image classification only needs to discriminate the basic-level category, such as bird or car. The difference between basic-level and fine-grained image classification is shown as Fig. \ref{fine-grained}.
Fine-grained image classification is a highly important task with wide applications, such as automatic driving, biological conservation and cancer detection.
Fig. \ref{interintra} shows the large variance in the same subcategory and small variance among different subcategories, and it is extremely hard for human beings to recognize hundreds of subcategories, such as 200 bird subcategories or 196 car subcategories. Due to small variance in object appearances, subtle and local differences are the key points for fine-grained image classification, such as the color of back, the shape of bill and the texture of feather for bird. 
Since these subtle and local differences locate at the discriminative objects and parts, most existing methods \cite{zhang2014part,zhang2016picking,zhang2016fused} generally follow the strategy of locating the objects or parts in the image and then discriminating which subcategory the image belongs to.

To localize the discriminative objects and parts, generating image patches with high objectness by a bottom-up process is generally first performed, meaning that the generated patches contain the discriminative object or parts.
Selective search \cite{uijlings2013selective} is an unsupervised method that can generate thousands of such image patches, which is extensively used in recent works \cite{girshick2014rich,zhang2014part,zhang2016picking}.
Since the bottom-up process has high recall but low precision, it is indispensable to remove the noisy image patches and retain those containing the object or discriminative parts, which can be achieved through top-down attention model.
In the context of fine-grained image classification, finding the objects and discriminative parts can be regarded as a two-level attention process, where one is object-level and the other is part-level.
An intuitive idea is to use object annotation (i.e. bounding box of object) for object-level attention and part annotations (i.e. part locations) for part-level attention. 
Most existing methods \cite{zhang2014part,branson2014bird,krause2015fine,zhou2016fine} rely on the object or part annotations for finding the object or discriminative parts, but such labeling is heavily labor consuming. This is the \emph{first limitation}.
\par
For addressing the above problem, researchers begin focusing on how to achieve promising performance under the weakly supervised setting that neither object nor part annotations are used in both training and testing phases. 
Zhang et al. \cite{zhang2016weakly} propose to select the discriminative parts through exploiting the useful information in part clusters. Zhang et al. \cite{zhang2016picking} propose an automatic fine-grained image classification method, incorporating deep convolutional filters for both part selection and description. However, 
when they select the discriminative parts, the spatial relationships between the object and its parts as well as among these parts are ignored, but both of them are highly helpful for finding the discriminative parts. 
This causes the selected parts: (1) have large areas of background noise and small areas of object, (2) have large overlap with each other which leads to redundant information.
This is the \emph{second limitation}.

\par 
For addressing the above two limitations, this paper proposes the object-part attention model (OPAM) for weakly supervised fine-grained image classification. Its main novelties and contributions can be summarized as follows:

\begin{itemize}
\item
\emph{\textbf{Object-Part Attention Model.}} \ Most existing works rely on object or part annotations \cite{zhang2014part,krause2015fine,zhou2016fine}, 
while labeling is heavily labor consuming. 
For addressing this important problem, we propose the object-part attention model for weakly supervised fine-grained image classification to avoid using the object and part annotations and march toward practical applications. It integrates two level attentions: 
(1) \emph{\textbf{Object-level attention model}} utilizes the global average pooling in CNN to extract the saliency map for localizing objects of images, which is to learn object features.
(2) \emph{\textbf{Part-level attention model}} first selects the discriminative parts and then aligns the parts based on the cluster patterns of neural network, which is to learn subtle and local features. 
The object-level attention model focuses on the representative object appearance, and the part-level attention model focuses on the distinguishing specific differences of parts among subcategories. 
Both of them are jointly employed to boost the multi-view and multi-scale feature learning, and enhance their mutual promotions to achieve good performance for fine-grained image classification.
\item
\emph{\textbf{Object-Part Spatial Constraint Model.}} \ Most existing weakly supervised methods \cite{zhang2016picking,zhang2016weakly} ignore the spatial relationships between the object and its parts as well as among these parts, both of which are highly helpful for discriminative part selection. For addressing this problem, we propose the part selection approach driven by object-part spatial constraint model, which 
combines two types of spatial constraints: (1) \emph{\textbf{Object spatial constraint}} enforces that the selected parts are located in the object region and highly representative. (2) \emph{\textbf{Part spatial constraint}} reduces the overlaps among parts and highlights the saliency of parts, which eliminates the redundancy and enhances the discrimination of selected parts. Combination of the two spatial constraints not only significantly promotes discriminative part selection by exploiting subtle and local distinction, but also achieves a notable improvement on fine-grained image classification. 
\end{itemize}

\par
Our previous conference paper \cite{xiao2015application} integrates two level attentions: object-level attention selects image patches relevant to the object, and part-level attention selects discriminative parts, which is the first work to classify fine-grained images without using object and part annotations in both training and testing phases, and achieves promising results \cite{zhang2016weakly}. 
In this paper, our OPAM approach further exploits the two level attentions to localize not only the discriminative parts but also the objects, and employs the object-part spatial constraint model to eliminate redundancy as well as highlight discrimination of the selected parts: 
\emph{\textbf{For object-level attention}}, we further propose \emph{an automatic object localization approach via saliency extraction} to focus on the representative object feature for better classification performance. It utilizes the global average pooling in CNN for localizing  objects of images, rather than only selecting the image patches relevant to object that have large areas of background noise or do not contain the whole object in image like \cite{xiao2015application}. 
\emph{\textbf{For part-level attention}}, we further propose \emph{a part selection approach driven by object-part spatial constraint model} to exploit the subtle and local differences among subcategories. It considers the spatial relationships between object and its parts as well as among these parts, thus avoids the problem of generating large areas of background noise and large overlaps among selected parts like \cite{xiao2015application}. 
Compared with more than 10 state-of-the-art methods on 4 widely-used datasets, the effectiveness of our OPAM approach is verified by the comprehensive experimental results.  
\par
The rest of this paper is organized as follows: Section \uppercase\expandafter{\romannumeral2} briefly reviews related works on fine-grained image classification. Section \uppercase\expandafter{\romannumeral3} presents our proposed OPAM approach, and Section \uppercase\expandafter{\romannumeral4} introduces the experiments as well as the result analyses. Finally Section \uppercase\expandafter{\romannumeral5} concludes this paper.

\section{Related Work}
Most traditional methods for fine-grained image classification follow the strategy of extracting basic low-level descriptors like SIFT \cite{lowe2004distinctive}, and then generating Bag-of-Words for image representation \cite{xie2014spatial,gao2014learning}. However, the performance of these methods is limited by the handcrafted features. Deep learning has shown its strong power in feature learning, and achieved great progresses in fine-grained image classification \cite{zhang2013deformable,branson2014bird,zhang2016fused,zhang2016picking,simon2015neural,jaderberg2015spatial,huang2016part,zhang2016spda,zhang2014part,xiao2015application,wang2015multiple,lin2015bilinear}. 
These methods can be divided into three groups \cite{zhaosurvey}: 
ensemble of networks based methods, visual attention based methods and part detection based methods. 

\subsection{Ensemble of Networks Based Methods}
Ensemble of networks based methods are proposed to utilize multiple neural networks to learn different representations of image for better classification performance. 
Each subcategory has an implied hierarchy of labels in its ontology tree. For example, Picoides Pubescens, which is the label in species level, has the label in genus level as Picoides and the family level as Picidae.
Wang et al. \cite{wang2015multiple} first leverage the labels of multiple levels to train a series of CNNs at each level, which focuses on different regions of interest in images. Different features are extracted by different level CNNs, and combined to encode informative and discriminative features. Finally, a linear SVM is trained to learn weights for the final classification. However, the external labels of ontology tree are necessary for the method of \cite{wang2015multiple}. Lin et al. \cite{lin2015bilinear} propose a bilinear CNN model, which is an end-to-end system jointly combining two CNNs, each of which is adopted as a feature extractor. The extracted features from two CNNs at each location of image are multiplied by outer product operation, and then pooled to generate an image descriptor. Finally, softmax is conducted for final prediction. Despite achieving promising results, these methods are still limited by the lack of ability to be spatially invariant to the input image. Therefore, Jaderberg et al. \cite{jaderberg2015spatial} propose a learnable network, called spatial transformer, which consists of three parts: localization network, grid generator and sampler. Four spatial transformers in parallel are performed on images, and capture the discriminative parts to pass to the part description subnets. Finally, softmax is conducted on the concatenated part descriptor for final prediction.

\subsection{Visual Attention Based Methods}
Due to attention system, humans focus on the discriminative regions of an image dynamically, rather than receiving and dealing with the information of entire image directly. This natural advantage makes the attention mechanism widely used in fine-grained image classification. 
Inspired by the way how humans perform visual sequence recognition, Sermanet et al. \cite{sermanet2014attention} propose the attention for fine-grained categorization (AFGC) system. 
First, they process a multi-resolution crop on the input image, where each crop is called a glimpse. Then they use the information of glimpses to output the next location and the next object via a deep recurrent neural network at each step. The final prediction is computed through the sequence of glimpses. Recently, fully convolutional neural network is used to learn the saliency of an image for finding the discriminative regions \cite{zhou2016cvpr}. Liu et al. \cite{liu2016fully} use the fully convolutional attention to localize multiple parts to get better classification performance. 
Xie et al. \cite{xie2016interactive} propose a novel algorithm, called InterActive, which computes the activeness (attention) of neurons and network connections, carrying high-level context as well as improving the descriptive power of low-level and mid-level neurons, thus achieves good performance on image classification. 
Zhou et al. \cite{zhou2016cvpr} use global average pooling (GAP) in CNN to generate the saliency map for each image. Based on the saliency map, the discriminative region can be found. Furthermore, a diversified visual attention network (DVAN) \cite{zhao2016diversified} is proposed to pursue the diversity of attention as well as gather discriminative information.
In this paper, our OPAM approach integrates two level attention models: object-level attention model focuses on the representative object appearance, and part-level attention model focuses on the discriminative parts. Both of them are jointly employed to learn multi-view and multi-scale features to enhance their mutual promotions.

\begin{figure*}[!ht]
  \begin{center}\includegraphics[width=1\linewidth]{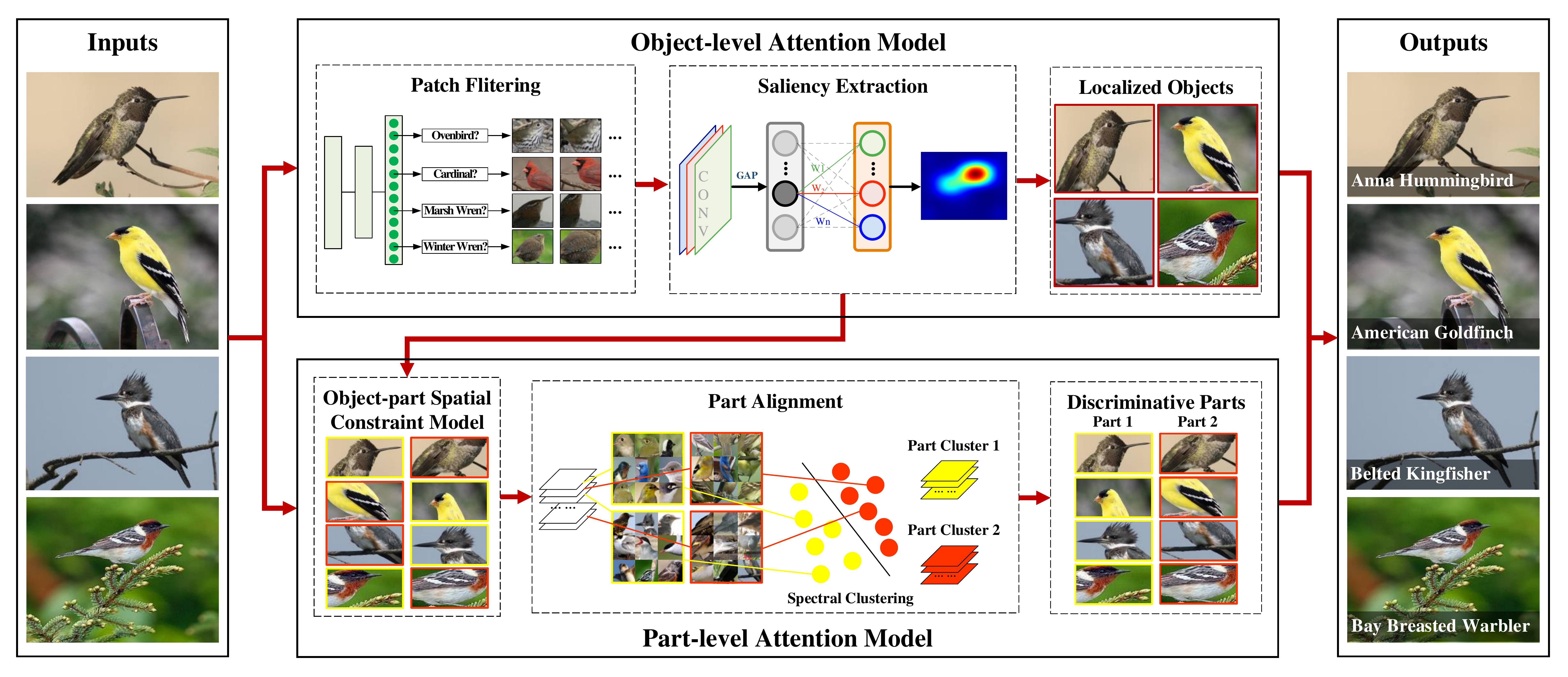}
  \caption{An overview of our OPAM approach. The object-level attention model is to localize object for learning object features. The part-level attention model is to select the discriminative parts for exploiting the subtle and local features. The outputs show the predicted subcategories.}
  \label{framework}
  \end{center}
\end{figure*}

\subsection{Part Detection Based Methods}
In fine-grained image classification, subtle and local differences generally locate at discriminative parts of object, so the discriminative part detection is crucial for fine-grained image classification. 
Girshick et al. \cite{girshick2014rich} propose a popular detection method, R-CNN, which first generates thousands of candidate image patches for each image via the bottom-up process \cite{uijlings2013selective}, and then selects the image patches with high classification scores as detection results. Zhang et al. \cite{zhang2014part} utilize R-CNN with a geometric prior to detect discriminative parts for fine-grained image classification, and then train a classifier on the features of detected parts for final categorization. They use both the object and part annotations. 

Recently, researchers begin focusing on how to detect the discriminative parts under the weakly supervised setting, which means neither object nor part annotations are used in both training and testing phases. 
Simon and Rodner \cite{simon2015neural} propose a constellation model to localize parts of object, leveraging CNN to find the constellations of neural activation patterns. First, neural activation maps are computed as part detectors by using the outputs of a middle layer of CNN. Second, a part model is estimated by selecting part detectors via constellation model. Finally, the part model is used to extract features for classification. 
Zhang et al. \cite{zhang2016picking} propose an automatic fine-grained image classification method, incorporating deep convolutional filters for both part selection and description. They combine two steps of deep filter response picking: The first step picks the discriminative filters that significantly respond to specific parts in image. The second step picks the salient regions and generates features with spatially weighted Fisher Vector based on the saliency map for classification.
Zhang et al. \cite{zhang2016weakly} propose to select the discriminative parts through exploiting the useful information in part clusters. 
In our OPAM approach, we first propose an object-part spatial constraint model to select discriminative parts, which considers the spatial relationships between object and its parts as well as among these parts, and then utilizes the cluster patterns of neural network to align the parts with the same semantic meaning together for improving the classification performance.

\section{Our OPAM Approach}
Our approach is based on an intuitive idea: fine-grained image classification generally first localizes the object (object-level attention) and then discriminative parts (part-level attention). For example, recognizing an image which contains a Field Sparrow follows the processes of first finding a bird, and then focusing on the discriminative parts that distinguish it from other bird subcategories. 
We propose the \emph{object-part attention model} for weakly supervised fine-grained image classification, which uses neither object nor part annotations in both training and testing phases, and only uses the image-level subcategory labels. 
As shown in Fig. \ref{framework}, our OPAM approach first localizes objects of images through object-level attention model for learning object features, and then selects the discriminative parts through part-level attention model for learning the subtle and local features. In the following subsections, the object-level and part-level attention models are presented respectively. 

\subsection{Object-level Attention Model}
Most existing weakly supervised works \cite{zhang2016picking,zhang2016weakly,simon2015neural} devote to the discriminative part selection, but ignore the object localization, which can remove the influence of background noise in image to learn meaningful and representative object features. Although some methods consider both object localization and part selection, they rely on the object and part annotations \cite{zhang2014part,zhang2013deformable}.
For addressing this important problem, we propose an object-level attention model based on the saliency extraction for localizing the objects of images automatically only with image-level subcategory labels, without any object or part annotations. The model consists of two components: patch filtering and saliency extraction. The first component is to filter out the noisy image patches and retain those relevant to the  object for training a CNN called \emph{ClassNet}, to learn multi-view and multi-scale features for the specific subcategory. 
The second component is to extract the saliency map via global average pooling in CNN for localizing the objects of images.

\begin{figure*}[!ht]
  \begin{center}\includegraphics[width=1\linewidth]{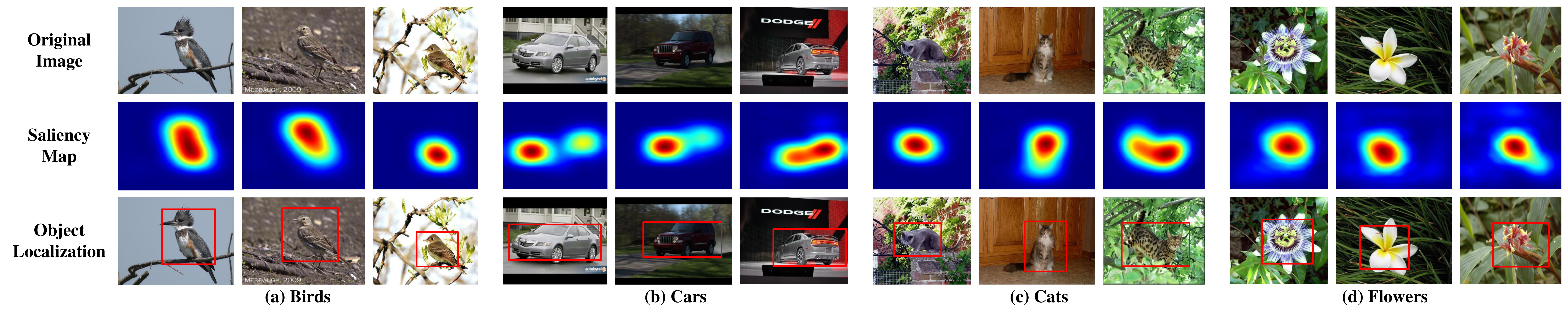}
  \caption{Some results of saliency extraction by our OPAM approach. The first row shows the original images and the second row shows the saliency maps of original images. The object localization results are shown in the third row, in which the red rectangles represent the bounding boxes automatically produced by saliency extraction. The images in (a) Birds, (b) Cars, (c) Cats and (d) Flowers are from CUB-200-2011 \cite{wah2011caltech}, Cars-196 \cite{krause20133d}, Oxford-IIIT Pet \cite{parkhi2012cats} and Oxford-Flower-102 \cite{Nilsback2008Automated} datasets respectively.}
  \label{saliencyextraction}
  \end{center}
\end{figure*}
\subsubsection{Patch Filtering}
A large amount of training data is significant for the performance of CNN, so we first focus on how to expand the training data. The bottom-up process can generate thousands of candidate image patches by grouping pixels into regions that may contain the object. 
These image patches can be used as the expansion of training data due to their relevances to the object. Therefore, selective search \cite{uijlings2013selective} is adopted to generate candidate image patches for a given image, which is an unsupervised and widely-used bottom-up process method. These candidate image patches provide multiple views and scales of original image, which benefit for training an effective CNN to achieve better fine-grained image classification accuracy. However, these patches can not be directly used due to the high recall but low precision, which means some noises exist. The object-level attention model is highly helpful for selecting the patches relevant to the object.
\par
We remove the noisy patches and select relevant patches through a CNN, called \emph{FilterNet}, which is pre-trained on the ImageNet 1K dataset \cite{imagenet_cvpr09}, and then fine-tuned on the training data. We define the activation of neuron in softmax layer belonging to the subcategory of input image as the selection confidence score, and then a threshold is set to decide whether the given candidate image patch should be selected or not. 
Then we obtain the image patches relevant to the object with multiple views and scales.
The expansion of training data improves the training effect of \emph{ClassNet}, which has two aspects of benefits for our OPAM approach: (1) \emph{ClassNet} is an effective fine-grained image classifier itself. (2) Its internal features are significantly helpful to build part clusters for aligning the parts with the same semantic meaning together, which will be described latter in Subsection B. It is noted that the patch filtering is performed only in the training phase and only uses image-level subcategory labels.

\subsubsection{Saliency Extraction}
In this stage, CAM \cite{zhou2016cvpr} is adopted to obtain the saliency map $M_c$ of an image for subcategory $c$ to localize the object. The saliency map indicates the representative regions used by the CNN to identify the subcategory of image, as shown in the second row of Fig. \ref{saliencyextraction}. 
Then object regions of images, as shown in the third row of Fig. \ref{saliencyextraction}, are obtained by performing binarization and connectivity area extraction on the saliency maps.

Given an image $I$, the activation of neuron $u$ in the last convolutional layer at spatial location $(x,y)$ is defined as $f_u(x,y)$, and $w_u^c$ defines the weight corresponding to subcategory $c$ for neuron $u$. The saliency value at spatial location $(x,y)$ for subcategory $c$ is computed as follows:
\begin{gather}
M_c(x,y) = \sum \limits_u w_u^c f_u(x,y)
\end{gather}
where $M_c(x,y)$ directly indicates the importance of activation at spatial location $(x,y)$ for classifying an image into subcategory $c$. 
Instead of using the image-level subcategory labels, we use the prediction result as the subcategory $c$ in saliency extraction for each image.
Through object-level attention model, we localize objects in the images to train a CNN called \emph{ObjectNet} for obtaining the prediction of object-level attention.

\begin{figure*}[!ht]
  \begin{center}\includegraphics[width=0.8\linewidth]{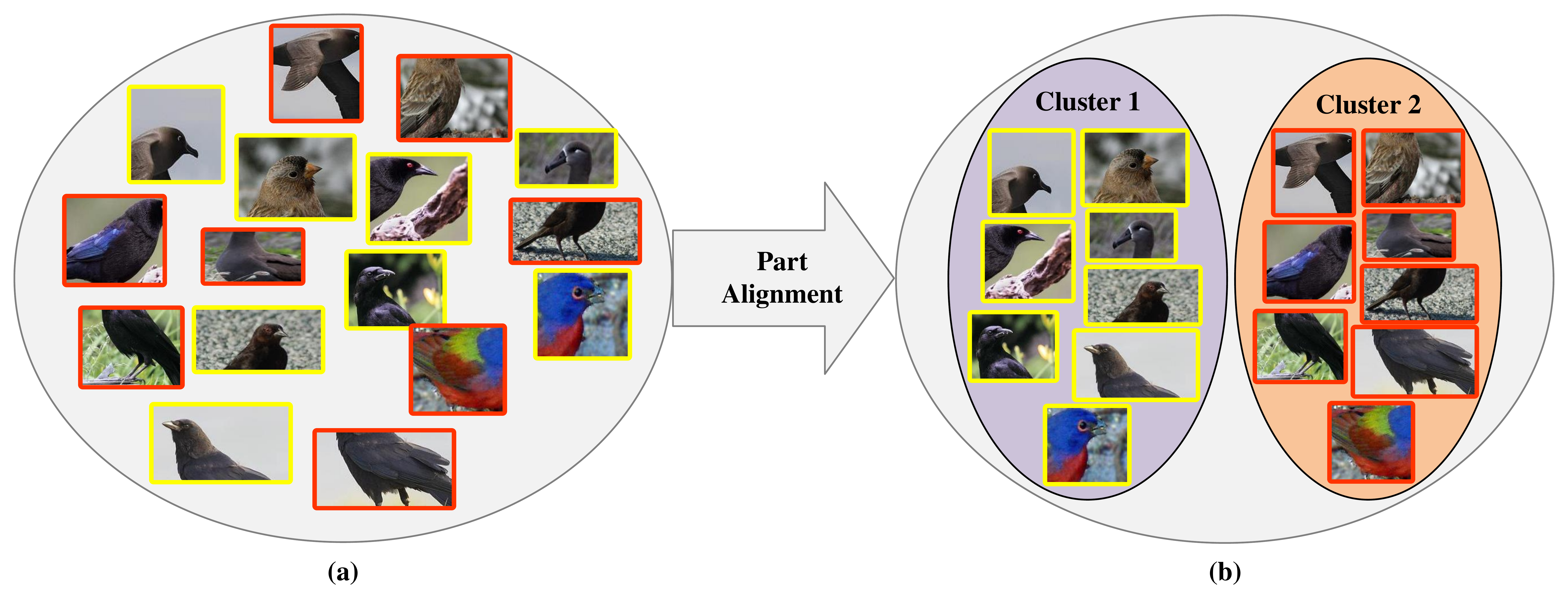}
  \caption{Some results of part alignment in our OPAM approach. (a) shows the image patches which are selected through object-part spatial constraint model, and (b) shows that the image patches are aligned into clusters via part clusters.}
  \label{partalign}
  \end{center}
\end{figure*}

\subsection{Part-level Attention Model}
Since the discriminative parts, such as head and body, are crucial for fine-grained image classification, previous works \cite{zhang2014part} select discriminative parts from the candidate image patches produced by the bottom-up process like selective search \cite{uijlings2013selective}. 
However, these works rely on the part annotations which are heavily labor consuming. Although some works begin to focus on finding the discriminative parts without using any part annotations \cite{xiao2015application,zhang2016picking}, 
they ignore the spatial relationships between the object and its parts as well as among these parts. 
Therefore, we propose a new part selection approach driven by part-level attention for exploiting the subtle and local discrimination to distinguish the subcategories, which uses neither object nor part annotations. 
It consists of two components: object-part spatial constraint model and part alignment. The first is to select the discriminative parts, and the second is to align the selected parts into clusters by the semantic meaning.

\subsubsection{Object-part Spatial Constraint Model}
We obtain object regions of images through object-level attention model, and then employ object-part spatial constraint model to select the discriminative parts from the candidate image patches produced by the bottom-up process. Two spatial constraints are jointly considered: \emph{object spatial constraint} defines the spatial relationship between object and its parts, and \emph{part spatial constraint} defines the spatial relationship among these parts. 
For a given image $I$, its saliency map $M$ and object region $b$ are obtained through object-level attention model. 
Then part selection is driven by object-part spatial constraint model as follows:

Let $\mathbb{P}$ denotes all the candidate image patches and $P = \{p_1, p_2, ..., p_n\}$ denotes $n$ parts selected from $\mathbb{P}$ as the discriminative parts for each given image. The object-part spatial constraint model considers the combination of two spatial constraints by solving the following optimization problem:
\begin{gather}
P^* = arg \max \limits_{\mathbb{P}} \Delta(P)
\end{gather}
where $\Delta(P)$ is defined as a scoring function over two spatial constraints as follows:
\begin{gather}
\Delta(P) = \Delta_{box}(P)\Delta_{part}(P)
\label{opeq}
\end{gather}
Eq. \ref{opeq} defines the proposed object-part spatial constraint, which ensures the representativeness and discrimination of the selected parts. It consists of two constraints: object spatial constraint $\Delta_{box}(P)$ and part spatial constraint $\Delta_{part}(P)$, which should be both satisfied by all the selected parts at the same time. For ensuring this, we choose product operation, not sum operation, as the work \cite{zhang2014part} which utilizes product operation to optimize two constraints.

\par
\textbf{\emph{Object spatial constraint.}}\ \ Ignoring the spatial relationship between the object and its parts causes that the selected parts may have large areas of background noise but small areas of discriminative region, which decreases the representativeness of selected parts. Since the discriminative parts are inside the object region, an intuitive spatial constraint function is defined as:
\begin{gather}
\Delta_{box}(P)=\prod_{i=1}^n f_b(p_i)
\label{objectspatialeq}
\end{gather}
where
\begin{gather}
f_b(p_i) = 
\begin{cases}
1, & IoU(p_i) > threshold \\
0, & otherwise
\end{cases} 
\end{gather}
and $IoU(p_i)$ defines the proportion of Intersection-over-Union (IoU) overlap of part region and object region. 
It is noted that the object region is obtained automatically through the object-level attention model, \emph{not provided by the object annotation}. 
Object spatial constraint aims to simultaneously restrain all the selected parts inside the object region. So product operation is utilized to ensure this, which is the same with the work \cite{zhang2014part}. That is to say, any part that does not satisfy object spatial constraint, e.g. its $IoU$ value equals $0$, will not be selected as a discriminative part.

\par
\textbf{\emph{Part spatial constraint.}}\ \ Ignoring the spatial relationship among these parts leads to the problem that the selected parts may have large overlap with each other, and some discriminative parts are ignored. The saliency map indicates the discrimination of image, and benefits for selecting discriminative parts. We jointly model saliency and the spatial relationship among parts as follows:
\begin{gather}
\Delta_{part}(P)=log(A_U-A_I-A_O) \nonumber \\
+ log(Mean(M_{A_U}))
\label{partpatialeq}
\end{gather}
where $A_U$ is the union area of $n$ parts, $A_I$ is the intersection area of $n$ parts, $A_O$ is the area outside the object region and $Mean(M_{A_U})$ is defined as follows:
\begin{figure}[!ht]
  \begin{center}\includegraphics[width=0.9\linewidth]{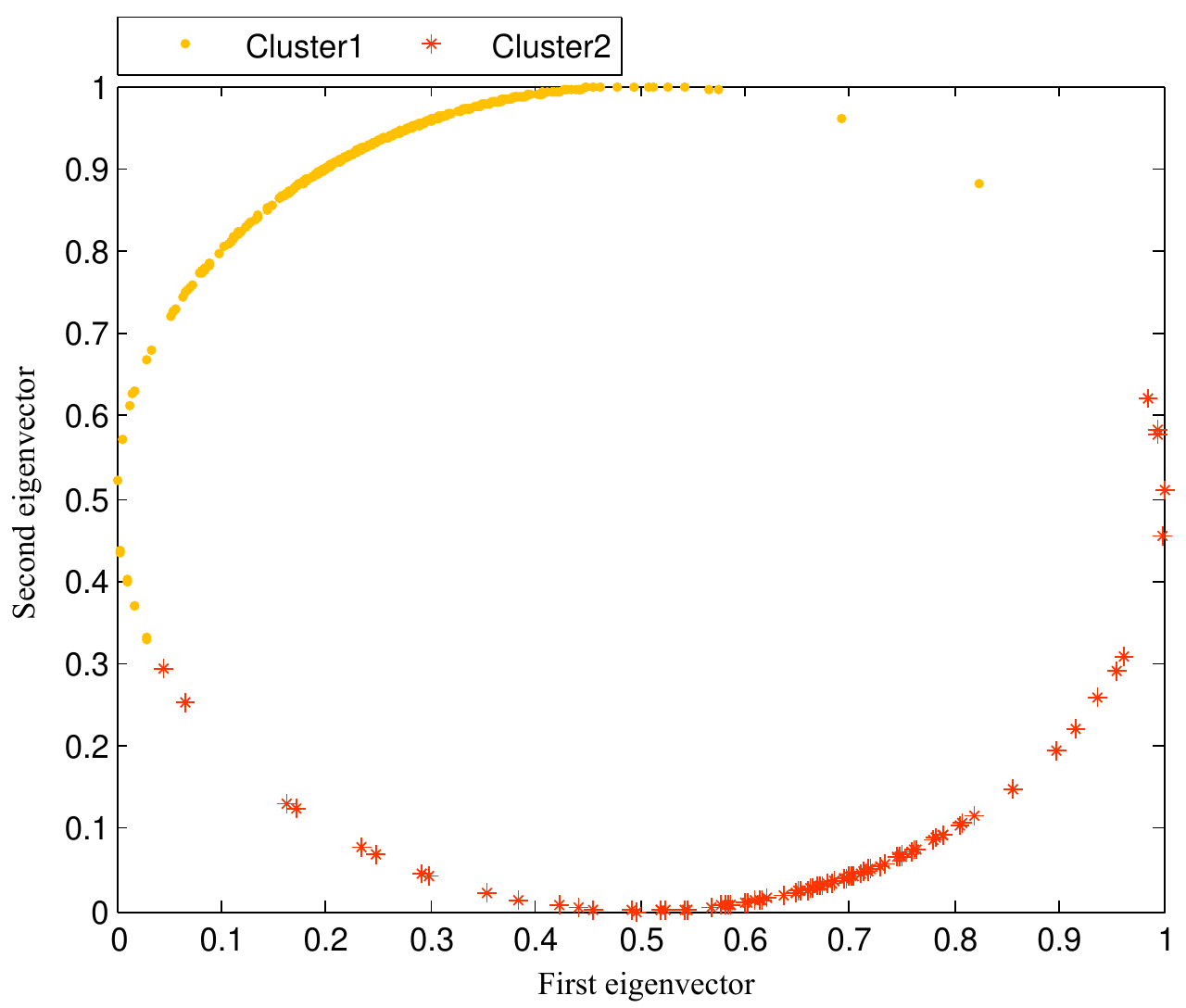}
  \caption{Illustration of spectral clustering. The coordinate values represent the two largest eigenvectors of similarity matrices among all neurons.}
  \label{spectralcluster}
  \end{center}
\end{figure}
\begin{gather}
Mean(M_{A_U}) = \frac{1}{|A_U|}\sum_{i,j}M_{ij}
\end{gather}
where pixel $(i,j)$ locates in the union area of parts, $M_{ij}$ refers to the saliency value of pixel $(i,j)$, and $|A_U|$ refers to the number of pixels that locate in the union area of $n$ parts. 
Part spatial constraint aims to select the most discriminative parts, which consists of two items: The first item aims to reduce the overlaps among selected parts, and is realized by $log(A_U-A_I-A_O)$, where $-A_I$ ensures the selected parts have the least overlap, and $-A_O$ ensures the selected parts have the largest areas inside the object region. The second item aims to maximize the saliency of selected parts, and is realized by $log(Mean(M_{A_U}))$, which denotes the average saliency value of all the pixels in the union area of selected parts. We hope both of the two items in Eq. \ref{partpatialeq} have the maximum values, so sum operation is adopted.

\subsubsection{Part Alignment}
The selected parts through object-part spatial constraint model are in disorder and not aligned by its semantic meaning, as shown in Fig. \ref{partalign}(a). These parts with different semantic meanings contribute to the final prediction differently, so an intuitive idea is to align the parts with the same semantic meaning together, as shown in Fig. \ref{partalign}(b).

\par
We are inspired by the fact that middle layers of \emph{ClassNet} show clustering patterns. For example, there are groups of neurons significantly responding to the head of bird, and others to the body of bird, despite the fact that they may correspond to different poses. 
So clustering is performed on the neurons of a middle layer in the \emph{ClassNet} to build the part clusters for aligning the selected parts.
We first compute the similarity matrix $S$, where $S(i,j)$ denotes the cosine similarity of weights between two mid-layer neurons $u_i$ and $u_j$, and then perform spectral clustering on the similarity matrix $S$ to partition the mid-layer neurons into $m$ groups. 
In the experiments, neurons are picked  from the penultimate convolutional layer with $m$ set as $2$, as shown in Fig. \ref{spectralcluster}, where the coordinate values represent the two largest eigenvectors of similarity matrices among all neurons, as the work \cite{nadler2006diffusion}.

Then we use the part clusters to align the selected parts as follows: (1) Warp the images of selected parts to the size of receptive field on input image of neuron in penultimate convolutional layer. (2) Feed forward the selected parts to the penultimate convolutional layer to produce an activation score for each neuron. (3) Sum up the scores of neurons in one cluster to get cluster score. (4) Align the selected parts to the cluster with highest cluster score, which is formulated as follows: 
For a given image, $n$ discriminative parts $P=\{p_1, p_2, ..., p_n\}$ are obtained by object-part spatial constraint model, and then part alignment is performed on these parts with $m$ part clusters $L=\{l_1, l_2, ..., l_m\}$ as Algorithm \ref{algopartalign}.

The choice of middle layer has important influence on the part alignment and classification performance. We follow standard practice and withhold a validation set of 10\% training data for grid search to determine which layer to choose. At last, we find the penultimate convolutional layer works better than others.
Through part-level attention model, we select the discriminative parts in images to train a CNN called \emph{PartNet} for obtaining the prediction of part-level attention.

\begin{figure}[!ht]
  \begin{center}\includegraphics[width=1\linewidth]{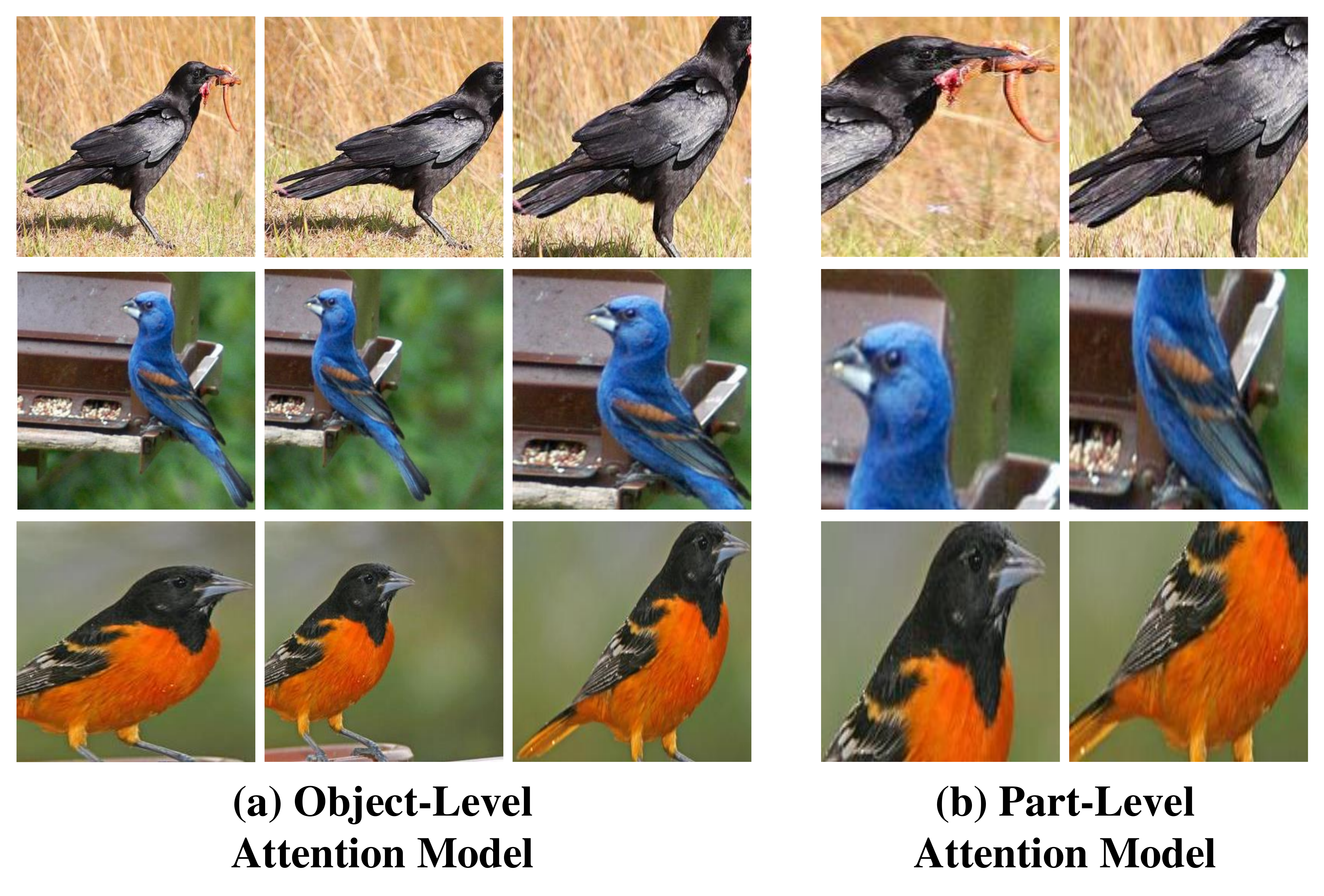}
  \caption{Some results of selected image patches by the object-level and part-level attention model respectively. Image patches selected by the object-level attention model focus on the whole objects, as shown in (a). Image patches selected by the part-level attention model focus on subtle and local features, as shown in (b).}
  \label{patch}
  \end{center}
\end{figure}

\begin{algorithm}
  \caption{Part Alignment}
  \label{algopartalign}
  \begin{algorithmic}[1]
    \REQUIRE  
    The $i$th selected part $p_i$; The part clusters $L = \{l_1, l_2, ..., l_m\}$; And the number of neurons in penultimate convolutional layer $d$.
    \ENSURE The cluster that $p_i$ is aligned into $l_{c}$.
    \STATE Set $score_{k} = 0; k = 1,...,m$.
    \STATE Warp $p_i$ to the size of receptive field on input image of neuron in penultimate convolutional layer.
    \STATE Perform a feed-forward pass to compute $p_i$'s activations $F_i = \{f_{i1}, f_{i2}, ..., f_{id}\}$.
    \FOR{$k = 1,...,m; j=1,...,d$}
    \IF {$j$th neuron belongs to cluster $l_{k}$}
    \STATE $score_{k} = score_{k} + f_{ij}$.
    \ENDIF
    \ENDFOR
    \STATE $c = \arg \max \limits_{k} score_{k}$.
    \RETURN $l_{c}$.
  \end{algorithmic}
\end{algorithm}

\subsection{Final Prediction}
For better classification performance, we fine-tune \emph{ClassNet} with the localized object and the discriminative parts to get two classifiers, called \emph{ObjectNet} and \emph{PartNet} respectively. \emph{ClassNet}, \emph{ObjectNet} and \emph{PartNet} are all fine-grained image classifiers: \emph{ClassNet} for original images, \emph{ObjectNet} for objects and \emph{PartNet} for selected discriminative parts. However, their impacts and strengths are different, primarily because they focus on the different natures of image.

Object-level attention model first drives \emph{FilterNet} to select image patches with multiple views and scales that are relevant to the object, as shown in Fig. \ref{patch} (a). These image patches drive \emph{ClassNet} to learn more representative features and localize the object region through saliency extraction. Part-level attention model selects discriminative parts which contain subtle and local features, as shown in Fig. \ref{patch} (b).
The different level focuses (i.e. original image, object of original image, and parts of original image) have different representations and are complementary to improve the prediction. Finally, we merge the prediction results of the three different levels by using the following equation:
\begin{gather} 
final\_score= \alpha * original\_score + \beta * object\_score \nonumber \\
+\gamma*part\_score
\label{predictioneq}
\end{gather}
where $original\_score, object\_score$ and $part\_score$ are the softmax values of \emph{ClassNet}, \emph{ObjectNet} and \emph{PartsNet} respectively, and $\alpha, \beta$ and $\gamma$ are selected by using the  $k$-fold cross-validation method \cite{kohavi1995study}. The subcategory with the highest $final\_score$ is chosen as the final prediction result.

\section{Experiments}
We conduct experiments on 4 widely-used datasets for fine-grained image classification: CUB-200-2011, Cars-196, Oxford-IIIT Pet and Oxford-Flower-102. Our proposed OPAM approach is compared with more than 10 state-of-the-art methods to verify its effectiveness.

\subsection{Datasets and evaluation metric}
Four datasets are adopted for the experiments:
\begin{itemize}
\item 
\textbf{CUB-200-2011} \cite{wah2011caltech}: It is the most widely-used dataset for fine-grained image classification, and contains 11788 images of 200 different bird subcategories, which is divided as follows: 5994 images for training and 5794 images for testing. For each subcategory, 30 images are selected for training and 11$\sim$30 images for testing, and each image has detailed annotations: a subcategory label, a bounding box of object, 15 part locations and 312 binary attributes. All attributes are visual in nature, pertaining to color, pattern, or shape of a particular part.
\item 
\textbf{Cars-196} \cite{krause20133d}: It contains 16185 images of 196 car subcategories, and is divided as follows: 8144 images for training and 8041 images for testing. For each subcategory, 24$\sim$84 images are selected for training and 24$\sim$83 images for testing. Each image is annotated with a subcategory label and a bounding box of object. 
\item 
\textbf{Oxford-IIIT Pet} \cite{parkhi2012cats}: It is a collection of 7349 images with 37 different pet subcategories, among which 12 are cat subcategories and 25 are dog subcategories. It is divided as follows: 3680 images for training and 3669 images for testing. For each subcategory, 93$\sim$100 images are selected for training and 88$\sim$100 images for testing. Each image is annotated with a subcategory label, a pixel level segmentation marking the body and a tight bounding box of head. 
\item
\textbf{Oxford-Flower-102} \cite{Nilsback2008Automated}: It has 8189 images of 102 subcategories belonging to flowers, 1020 for training, 1020 for validation and 6149 for testing. One image may contains several flowers. Each image is annotated with a subcategory label.
\end{itemize}


\emph{Accuracy} is adopted as the evaluation metric to comprehensively evaluate the classification performances of our OPAM approach and compared methods, which is widely used for evaluating the performance of fine-grained image classification \cite{zhang2016weakly,zhang2016picking,zhang2014part}, and is defined as follows:
\begin{gather}
Accuracy = \frac{R_a}{R}
\end{gather} 
where $R$ means the number of testing images and $R_a$ counts the number of images which are correctly classified.

\subsection{Details of the networks}
In the experiments, the widely-used CNN of VGGNet \cite{simonyan2014very} is adopted. It is noted that the CNN used in our proposed approach can be replaced with the other CNNs. In our approach, CNN has two different effects: localization and classification. Therefore, the architectures of CNNs are modified for different functions:

\subsubsection{Localization} In the object-level attention model, CNN is used to extract the saliency map of an image for object localization. Zhou et al. \cite{zhou2016cvpr} find that the accuracy of localization can be improved if the last convolutional layer before global average pooling has a higher spatial resolution, which is termed as the mapping resolution. In order to get a higher spatial resolution, the layers after conv5\_3 are removed, resulting in a mapping resolution of $14 \times 14$. Besides, a convolutional layer of size $3 \times 3$, stride $1$, pad $1$ with $1024$ neurons is added, followed by a global average pooling layer and a softmax layer. 
The modified VGGNet is pre-trained on the $1.3M$ training images of ImageNet 1K dataset \cite{imagenet_cvpr09}, and then fine-tuned on the fine-grained image classification dataset. The number of neurons in softmax layer is set as the number of subcategories.

\subsubsection{Classification} The CNN used in the experiments for classification is the VGGNet \cite{simonyan2014very} with batch normalization \cite{ioffe2015batch}. For the prediction results of original image, object and parts, the same CNN architecture is used but fine-tuned on different training data. For the prediction of original image, we fine-tune the CNN on the image patches selected through object-level attention model, as \emph{ClassNet}. For the predictions of object and part, we fine-tuned the CNNs on the images of objects and images of parts based on \emph{ClassNet} respectively, as \emph{ObjectNet} and \emph{PartNet}.  
Then we can get prediction results of the three different levels in Eq. \ref{predictioneq}. We follow the work \cite{zhang2014part} to select the $3$ parameters (i.e. $\alpha$, $\beta$ and $\gamma$) by $k$-fold cross-validation method \cite{kohavi1995study}. 
Considering that the scale of training dataset is small, we set $k$ as $3$ to ensure that each subset of the training dataset is not too small, which guarantees a better selection of parameters. We follow \cite{kohavi1995study} to randomly split the training dataset $D$ into $3$ mutually exclusive subsets $D_1,D_2,D_3$ of equal size. We conduct experiment $3$ times. For each time $t$, we train on $D\backslash D_t$ and test on $D_t$. For parameter selection, we traverse the value of each parameter from $0$ to $1$ by step of $0.1$. We select the parameters that obtain the highest classification accuracy. Finally, for CUB-200-2011, Cars-196, Oxford-IIT Pet and Oxford-Flower-102 datasets, $(\alpha, \beta, \gamma)$ are set as $(0.4, 0.4, 0.2)$, $(0.5, 0.3, 0.2)$, $(0.4, 0.4, 0.2)$ and $(0.4, 0.3, 0.3)$.

\begin{table*}[!ht]
  \centering
  \caption{Comparisons with State-of-the-art Methods on CUB-200-2011 dataset.}
  \label{cub}
  \begin{tabular} {|c|c|c|c|c|c|c|}
    \hline
    \multirow {2}{*}{Method} & \multicolumn{2}{c|}{Train Annotation} & \multicolumn{2}{c|}{Test Annotation} & \multirow {2}{*}{Accuracy (\%)} & \multirow {2}{*}{CNN Features}\\
    \cline{2-5}
    &Object & Parts & Object & Parts & &\\
    \hline
    \textbf{Our OPAM Approach} & & & & & {\textbf{85.83}} & VGGNet \\
    \hline
    FOAF \cite{zhang2016fused} & & & & & 84.63 & VGGNet \\
    PD \cite{zhang2016picking}& & & & & 84.54 & VGGNet \\
    STN \cite{jaderberg2015spatial}& & & & & 84.10 & GoogleNet \\
    Bilinear-CNN \cite{lin2015bilinear}&  & &  & & 84.10 & VGGNet\&VGG-M \\
    Multi-grained \cite{wang2015multiple} & & & & & 81.70 & VGGNet \\
    NAC \cite{simon2015neural} & & & & & 81.01 & VGGNet \\
    PIR \cite{zhang2016weakly}& & & & & 79.34 & VGGNet \\
    TL Atten \cite{xiao2015application} & & & & & 77.90 & VGGNet \\
    MIL \cite{xu2017friend} & & & & & 77.40 & VGGNet \\
    VGG-BGLm \cite{zhou2016fine} & & & & & 75.90 & VGGNet \\
    InterActive \cite{xie2016interactive} & & & & & 75.62 & VGGNet \\
    Dense Graph Mining \cite{zhang2016detecting} & & & & & 60.19 &  \\
    \hline
    Coarse-to-Fine \cite{yao2016coarse}& $\surd$ & & & & 82.50 & VGGNet \\
    Coarse-to-Fine \cite{yao2016coarse}& $\surd$ & & $\surd$ & & 82.90 & VGGNet \\
    PG Alignment \cite{krause2015fine} & $\surd$ & & $\surd$ & & 82.80 & VGGNet \\
    VGG-BGLm \cite{zhou2016fine} & $\surd$ & & $\surd$ & & 80.40 & VGGNet \\
    Triplet-A (64) \cite{cui2015fine} & $\surd$ & & $\surd$ & & 80.70 & GoogleNet \\
    Triplet-M (64) \cite{cui2015fine} & $\surd$ & & $\surd$ & & 79.30 & GoogleNet \\
    \hline
    Webly-supervised \cite{xu2016webly} & $\surd$ & $\surd$ &  &  & 78.60 & AlexNet \\
    PN-CNN \cite{branson2014bird} & $\surd$ & $\surd$ &  &  & 75.70 & AlexNet \\
    Part-based R-CNN \cite{zhang2014part} & $\surd$ & $\surd$ &  & & 73.50 & AlexNet \\
    SPDA-CNN \cite{zhang2016spda} & $\surd$ & $\surd$ & $\surd$ &  & 85.14 & VGGNet \\
    Deep LAC \cite{lin2015deep} & $\surd$ & $\surd$ & $\surd$ &  & 84.10 & AlexNet \\
    SPDA-CNN \cite{zhang2016spda} & $\surd$ & $\surd$ & $\surd$ &  & 81.01 & AlexNet \\
    PS-CNN \cite{huang2016part}& $\surd$ & $\surd$ & $\surd$ &  & 76.20 & AlexNet \\
    PN-CNN \cite{branson2014bird} & $\surd$ & $\surd$ & $\surd$ & $\surd$ & 85.40 & AlexNet  \\
    Part-based R-CNN \cite{zhang2014part} & $\surd$ & $\surd$ & $\surd$ & $\surd$ & 76.37 & AlexNet \\
    POOF \cite{berg2013poof} & $\surd$ & $\surd$ & $\surd$ & $\surd$ & 73.30 &  \\
    HPM \cite{xie2013hierarchical} & $\surd$ & $\surd$ & $\surd$ & $\surd$ & 66.35 &  \\
    \hline
  \end{tabular}
\end{table*}

\begin{table*}[!ht]
  \centering
  \caption{Comparisons with State-of-the-art Methods on Cars-196 dataset.}
  \label{car}
  \begin{tabular} {|c|c|c|c|c|c|c|}
    \hline
    \multirow {2}{*}{Method} & \multicolumn{2}{c|}{Train Annotation} & \multicolumn{2}{c|}{Test Annotation} & \multirow {2}{*}{Accuracy (\%)} & \multirow {2}{*}{CNN Features}\\
    \cline{2-5}
    &Object & Parts & Object & Parts & &\\
    \hline
    \textbf{Our OPAM Approach} & & & & & {\textbf{92.19}} & VGGNet \\
    \hline
    Bilinear-CNN \cite{lin2015bilinear} & & & & & 91.30 & VGGNet\&VGG-M \\
    TL Atten \cite{xiao2015application}& & & & & 88.63 & VGGNet\\
    DVAN \cite{zhao2016diversified} & & & & & 87.10 & VGGNet \\
    FT-HAR-CNN \cite{xie2015hyper} & & & & & 86.30 & AlexNet \\
    HAR-CNN \cite{xie2015hyper} & & & & & 80.80 & AlexNet \\
    \hline
    PG Alignment \cite{krause2015fine}& $\surd$ & & & & 92.60 & VGGNet \\
    ELLF \cite{krause2014learning} & $\surd$ & & & & 73.90 & CNN \\
    R-CNN \cite{girshick2014rich} & $\surd$ & & & & 57.40 & AlexNet \\
    PG Alignment \cite{krause2015fine} & $\surd$ & & $\surd$ & & 92.80 & VGGNet \\
    BoT(CNN With Geo) \cite{wang2016mining}& $\surd$ & & $\surd$ & & 92.50 & VGGNet \\
    DPL-CNN \cite{wang2016weakly}& $\surd$ & & $\surd$ & & 92.30 & VGGNet \\
    VGG-BGLm \cite{zhou2016fine} & $\surd$ & & $\surd$ & & 90.50 & VGGNet \\
    LLC \cite{wang2010locality} & $\surd$ & & $\surd$ & & 69.50 &  \\
    BB-3D-G \cite{krause20133d} & $\surd$ & & $\surd$ & & 67.60 &  \\
    \hline
  \end{tabular}
\end{table*}

\begin{table*}[!ht]
  \centering
  \caption{Comparisons with State-of-the-art Methods on Oxford-IIIT Pet dataset.}
  \label{pet}
  \begin{tabular}{|p{6cm}<{\centering}|p{3cm}<{\centering}|p{2.5cm}<{\centering}|}
    \hline
    Method & Accuracy (\%) & CNN Features \\
    \hline
    \textbf{Our OPAM Approach} &{\textbf{93.81}} & VGGNet \\
    \hline
    InterActive \cite{xie2016interactive} & 93.45& VGGNet \\
    TL Atten \cite{xiao2015application} & 92.51 & VGGNet\\
    NAC \cite{simon2015neural}& 91.60 & VGGNet \\ 
    FOAF \cite{zhang2016fused}& 91.39 & VGGNet \\ 
    ONE+SVM \cite{xie2015image}& 90.03 & VGGNet \\ 
    Deep Optimized \cite{azizpour2015generic} & 88.10 & AlexNet \\ 
    NAC \cite{simon2015neural} & 85.20 & AlexNet \\ 
    MsML+ \cite{qian2015fine}& 81.18 & CNN \\
    MsML \cite{qian2015fine}& 80.45 & CNN \\
    Deep Standard \cite{azizpour2015generic} & 78.50 & AlexNet \\
    Shape+Appearance \cite{parkhi2012cats} & 56.68 &  \\ 
    Zernike+SCC \cite{iscen2015comparison}& 59.50 &  \\
    GMP+p \cite{murray2014generalized} & 56.80 &  \\
    GMP \cite{murray2014generalized} & 56.10 &  \\
    M-HMP \cite{bo2013multipath} & 53.40 &  \\
    Detection+Segmentation \cite{angelova2013efficient} & 54.30 & \\ 
    \hline
  \end{tabular}
\end{table*}

\begin{table*}
   \begin{center}
   \caption{Comparisons with state-of-the-art methods on Oxford-Flower-102 dataset.}
   \begin{tabular}{|p{6cm}<{\centering}|p{3cm}<{\centering}|p{2.5cm}<{\centering}|}
      \hline
      Method & Accuracy (\%) & CNN Features\\ 
      \hline
      \hline
      \textbf{Our OPAM Approach} & \textbf{97.10} & VGGNet\\ 
      \hline
      InterActive \cite{xie2016interactive} & 96.40 & VGGNet \\
      PBC \cite{huang2017pbc} & 96.10 & GoogleNet\\
      TL Atten \cite{xiao2015application} & 95.76 & VGGNet \\
      NAC \cite{simon2015neural}& 95.34 & VGGNet \\
      RIIR \cite{xie2017towards} & 94.01 & VGGNet  \\
      Deep Optimized \cite{azizpour2015generic} & 91.30 & AlexNet\\
      SDR \cite{azizpour2015generic}& 90.50 & AlexNet\\
      MML \cite{qian2015fine} & 89.45 & CNN \\
      CNN Feature \cite{sharif2014cnn} & 86.80 & CNN \\
      Generalized Max Pooling \cite{murray2014generalized} & 84.60  &\\  
      Efficient Object Detection \cite{plant} & 80.66 &\\
      \hline
   \end{tabular}
      \label{flower}
   \end{center}
\end{table*}

\subsection{Comparisons with the state-of-the-art methods}
This subsection presents the experimental results and analyses of our OPAM approach on 4 widely-used fine-grained image classification datasets as well as the state-of-the-art methods. 
Table \ref{cub} shows the comparison results on CUB-200-2011 dataset. The object, part annotations and CNN features used in these methods are listed for fair comparison. CNN models shown in the column of ``CNN Features'', such as “AlexNet”, “VGGNet” and “GoogleNet”, indicate which CNN model this method adopts to extract CNN features. If the column is empty, it means that the result of this method is produced by handcrafted feature like SIFT.

Early works \cite{berg2013poof,xie2013hierarchical,zhang2016detecting} choose SIFT \cite{lowe2004distinctive} as features, and the performances are limited and much lower than our OPAM approach no mater whether using the object and part annotations or not. 
Our approach is the best among all methods under the same setting that neither object nor part annotations are used in both training and testing phases, 
and obtains 1.20\% higher accuracy than the best compared result of FOAF \cite{zhang2016fused} (85.83\% vs. 84.63\%). It is noted that the CNN used in FOAF is pre-trained not only on ImageNet 1K dataset \cite{imagenet_cvpr09} but also on the dataset of PASCAL VOC \cite{azizpour2012object}, while our approach does not use the external dataset like PASCAL VOC. Compared with the second highest result of PD \cite{zhang2016picking}, our approach achieves 1.29\% higher accuracy (85.83\% vs. 84.54\%). 
Our OPAM approach improves 7.93\% than our previous conference paper \cite{xiao2015application}, and it verifies the effectiveness of further exploitation in our OPAM approach, which jointly integrates the object-level and part-level attention models to boost the multi-view and multi-scale feature learning and enhance their complementarity. Besides, our OPAM approach employs the object-part spatial constraint model to exploit the subtle and local discrimination for distinguishing the
subcategories. 

Our approach performs better than the methods which focus on the CNN architectures, such as STN \cite{jaderberg2015spatial} and Bilinear-CNN \cite{lin2015bilinear}. 
In STN, GoogleNet \cite{szegedy2015going} with batch normalization \cite{ioffe2015batch} is adopted to achieve the accuracy of 82.30\% by only fine-tuning on CUB-200-2011 dataset without any other processing. Two different CNNs are employed in Bilinear-CNN: VGGNet \cite{simonyan2014very} and VGG-M \cite{chatfield2014return}. The classification accuracies of the two methods are both 84.10\%, which are lower than our approach by 1.73\%. 

Furthermore, our approach outperforms the methods which use object annotation, such as Coarse-to-Fine \cite{yao2016coarse}, PG Alignment \cite{krause2015fine} and VGG-BGLm \cite{zhou2016fine}.  
Moreover, our approach outperforms methods that use both object and part annotations \cite{zhang2014part,zhang2016spda}. 
Neither object nor part annotations are used in our OPAM approach, which makes fine-grained image classification march toward practical application.

Besides, the results on Cars-196, Oxford-IIIT Pet and Oxford-Flower-102 datasets are shown in Tables \ref{car}, \ref{pet} and \ref{flower} respectively. The trends of results on these three datasets are similar as CUB-200-2011 dataset, our OPAM approach achieves the best results among state-of-the-art methods (92.19\%, 93.81\% and 97.10\% respectively) and 
brings 0.89\%, 0.36\% and 0.70\% improvements than the best results of compared methods respectively.

\subsection{Performances of components in our OPAM approach}
Detailed experiments are performed on our OPAM approach from the following three aspects:

\begin{table*}[!ht]
  \centering
  \caption{Performances of components in our OPAM approach on CUB-200-2011, Cars-196, Oxford-IIIT Pet and Oxford-Flower-102 datasets. 
  }
  \label{variants}
   \begin{tabular} {|c|c|c|c|c|}
    \hline
    \multirow {2}{*}{Method} & \multicolumn{4}{c|}{Accuracy (\%)} \\
    \cline{2-5}
        &CUB-200-2011&Cars-196&Oxford-IIIT Pet& Oxford-Flower-102\\
    \hline
    {\begin{tabular}{c} \textbf{Our OPAM Approach} \\ \textbf{(Original+Object-level+Part-level)} \end{tabular} } & \textbf{85.83} & \textbf{92.19} & \textbf{93.81} & \textbf{97.10}\\
    \hline
    Original & 80.82 & 86.79 & 88.14 & 94.70\\
    Object-level & 83.74 &  88.79 &  90.98 & 95.32 \\
    Part-level &  80.65 &  84.26 &  85.75 & 93.09  \\
    Original+Object-level & 84.79 & 91.15 & 92.20 & 96.55\\
    Original+Part-level & 84.41 & 91.06 & 91.82 & 96.23\\
    Object-level+Part-level & 84.73 & 89.69 & 91.50 & 95.66\\
    \hline
  \end{tabular}
\end{table*}

\begin{figure*}[!ht]
  \begin{center}\includegraphics[width=0.96\linewidth]{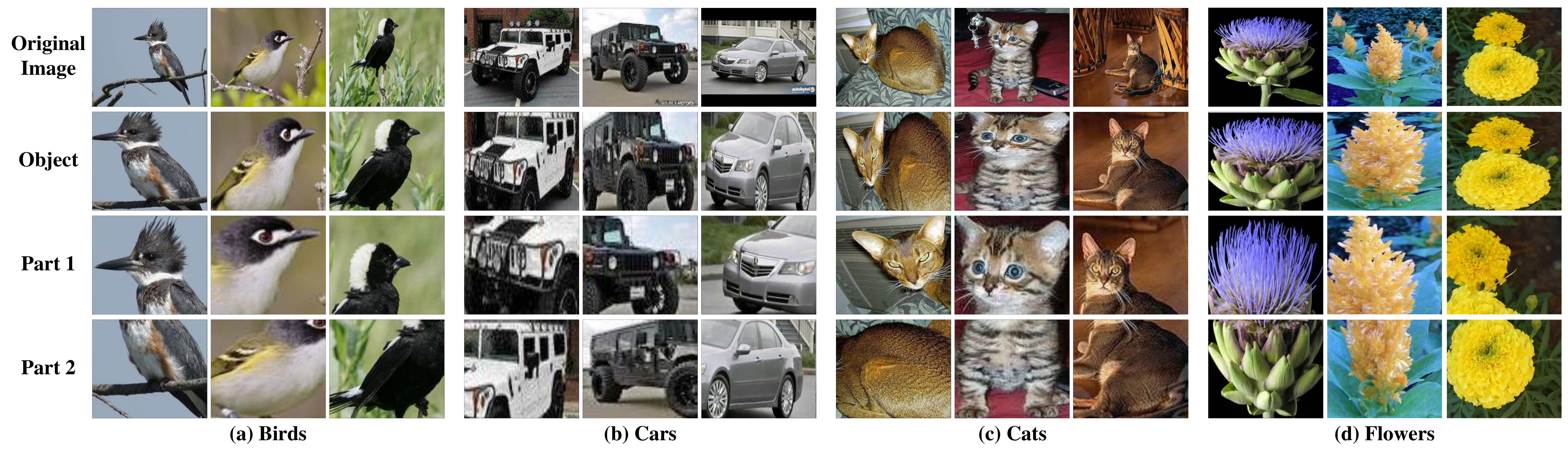}
  \caption{Some results of object localization and part selection. The first row denotes the original images, the second row denotes the localized objects of original images via object-level attention model, and the third and fourth rows denote the selected discriminative parts via part-level attention model. The images in (a) Birds, (b) Cars, (c) Cats and (d) Flowers are from CUB-200-2011 \cite{wah2011caltech}, Cars-196 \cite{krause20133d}, Oxford-IIIT Pet \cite{parkhi2012cats} and Oxford-Flower-102 \cite{Nilsback2008Automated} datasets respectively.}
  \label{partresult}
  \end{center}
\end{figure*}

\begin{figure}[!ht]
  \begin{center}\includegraphics[width=0.9\linewidth]{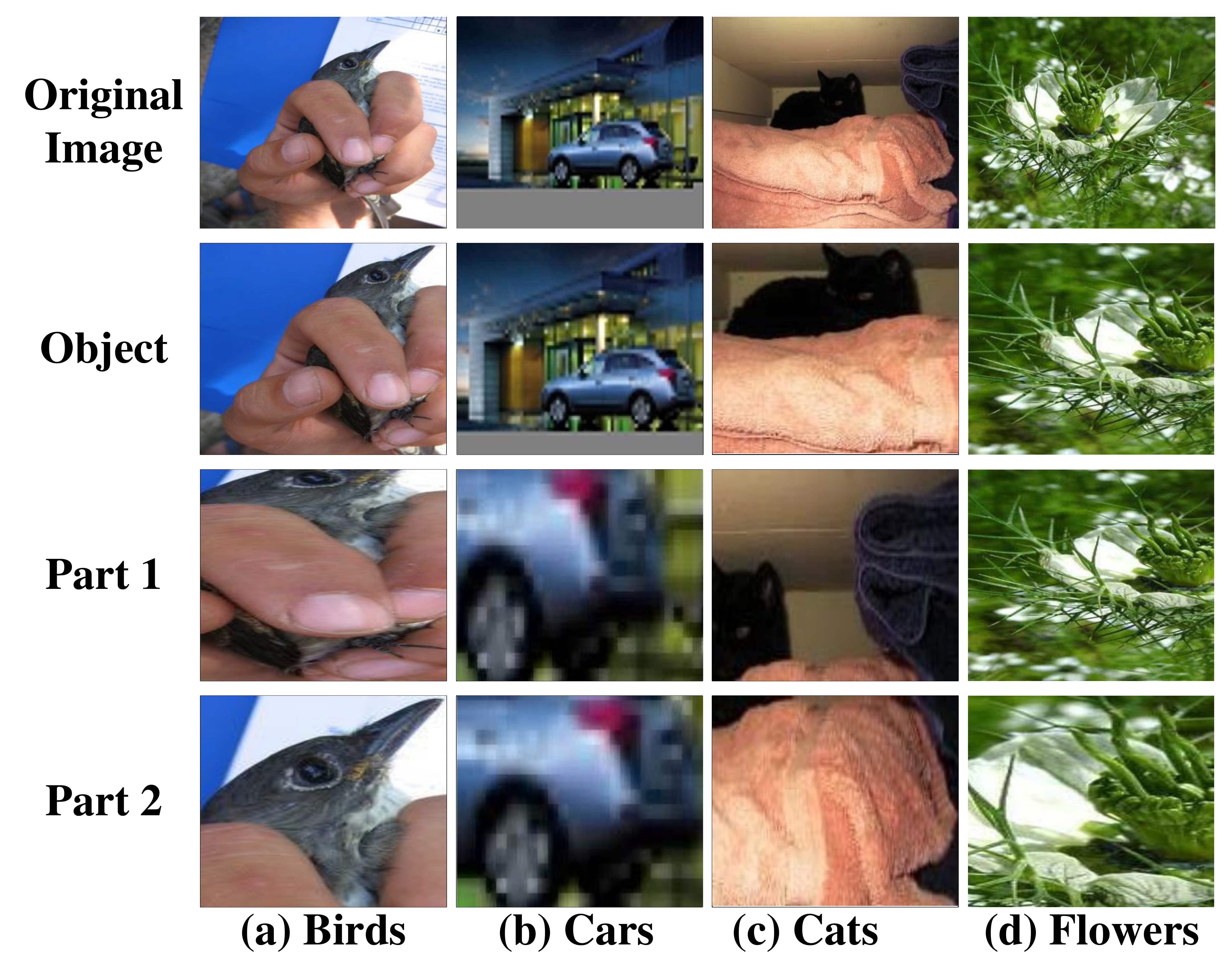}
  \caption{Some failure results of part selection. The images in (a) Birds, (b) Cars, (c) Cats and (d) Flowers are from CUB-200-2011 \cite{wah2011caltech}, Cars-196 \cite{krause20133d}， Oxford-IIIT Pet \cite{parkhi2012cats} and Oxford-Flower-102 \cite{Nilsback2008Automated} datasets respectively.}
  \label{errorpartresult}
  \end{center}
\end{figure}

\subsubsection{Effectivenesses of object-level attention and part-level attention models} 
In our OPAM approach, the final prediction score is generated by merging the prediction scores of three different images, i.e. original image, image of object and images of parts, which are denoted as ``Original'', ``Object-level'' and ``Part-level''. 
The effectivenesses of object-level and part-level attention models are verified in the following paragraphs. From Table \ref{variants}, Fig. \ref{partresult} and \ref{errorpartresult}, we can observe that:

\begin{itemize}
\item
Object-level attention model improves the classification accuracy via localizing objects of images for learning global features. Comparing with the result of ``Original'', it improves by 2.92\%, 2.00\%, 2.84\% and 0.62\% on four datasets respectively, and combining ``Object-level'' with ``Original'' improves even more, i.e. by 3.97\%, 4.36\%, 4.06\% and 1.85\% on four datasets respectively. The classification accuracy of part-level attention model is not higher than ``Original''. Fig. \ref{errorpartresult} shows some failure results of part selection. We conclude that our proposed part selection approach may fail in following two cases: 1) Objects are hard to be distinguished from the background. 2) Objects are in heavy occlusion. In these two cases, it is hard to localize the object accurately so the part selection fails, which is based on the object localization. The failure of part selection is the first reason of lower accuracy only with part. Another reason is that part-level attention focuses on the subtle and local features of object, containing less information than original image. However, despite these challenging cases, ``Part-level'' still achieves considerable classification accuracies, which is better than some state-of-the-art methods, such as \cite{zhou2016fine,xu2017friend}. Besides, it is complementary with original image and object, so their combination further boosts the classification accuracy and achieves the best result compared with state-of-the-art methods.


\item
Combining object-level and part-level attention models achieves more accurate results than only one level attention model, e.g. 84.73\% vs. 83.74\% and 80.65\% on CUB-200-2011 dataset. Combining the two level attention models with ``Original'' improves a lot than ``Original'', i.e. by 5.01\%, 5.40\%, 5.67\% and 2.4\% on the four datasets respectively. It shows the complementarity of object-level and part-level attention models in fine-grained image classification. The two level attention models have different but complementary focuses: the object-level attention model focuses on differences of representative object appearances, while the part-level attention model focuses on the subtle and local differences of discriminative parts among subcategories. Both of them are jointly employed to boost the multi-view and multi-scale feature learning and enhance their mutual promotions to achieve better performance for fine-grained image classification.

\begin{figure*}[!ht]
  \begin{center}\includegraphics[width=0.85\linewidth]{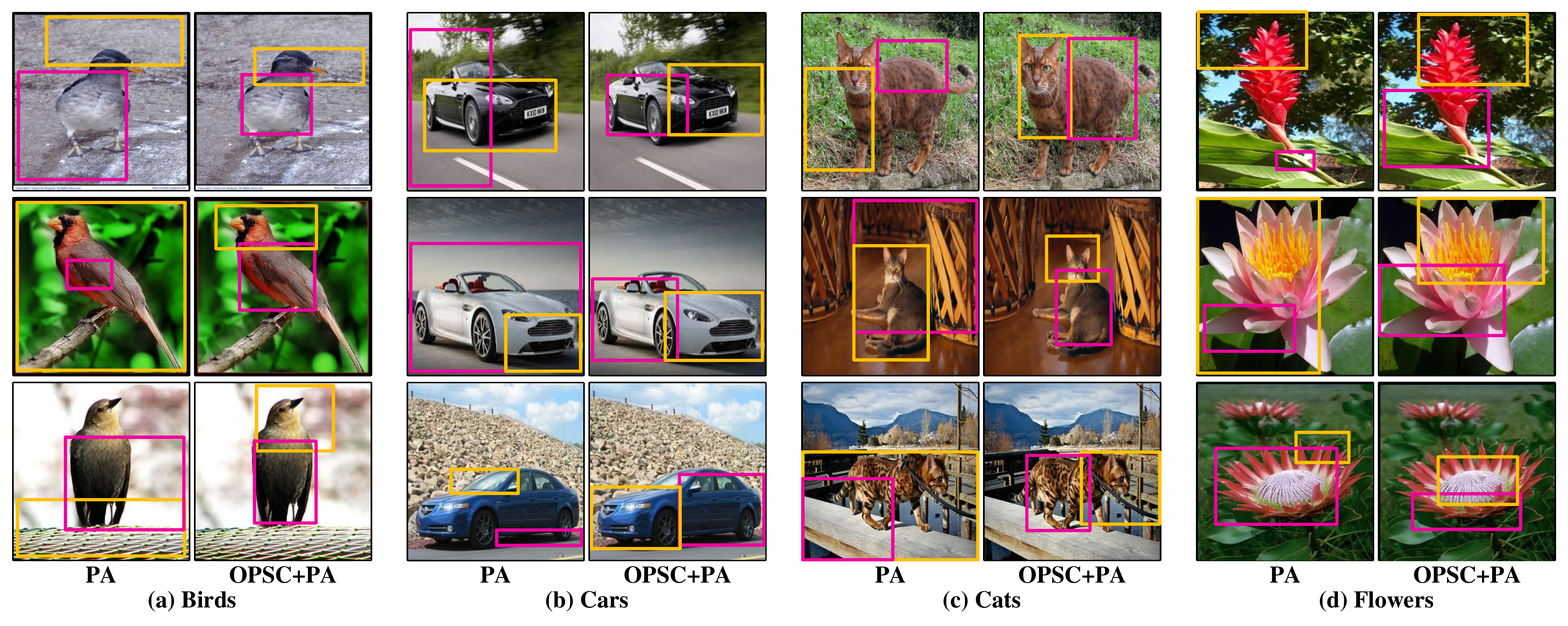}
  \caption{Examples of part selection from our previous conference paper \cite{xiao2015application} (left column) and our OPAM approach in this paper (right column). ``PA'' refers to part alignment which is adopted in our previous conference paper \cite{xiao2015application}, ``OPSC'' refers to object-part spatial constraint model, and ``OPSC+PA'' refers to combining the above two approaches, which is adopted in our OPAM approach. The yellow and orange rectangles denote the selected discriminative parts via the two approaches, which respond to the heads and bodies of objects. The images in (a) Birds, (b) Cars, (c) Cats and (d) Flowers are from CUB-200-2011 \cite{wah2011caltech}, Cars-196 \cite{krause20133d}, Oxford-IIIT Pet \cite{parkhi2012cats} and Oxford-Flower-102 \cite{Nilsback2008Automated} datasets respectively.}
  \label{cluster}
  \end{center}
\end{figure*}
\begin{table*}[!ht]
  \centering
  \caption{Performances of object-part spatial constraint model, part alignment and their combination.
  }
  \label{SCPA}
  \begin{tabular} {|p{2.6cm}<{\centering}|p{2.1cm}<{\centering}|p{2.1cm}<{\centering}|p{2.1cm}<{\centering}|p{2.5cm}<{\centering}|}
   \hline
    \multirow {2}{*}{Method} & \multicolumn{4}{c|}{Accuracy (\%)} \\
    \cline{2-5}
        & CUB-200-2011 & Cars-196 & Oxford-IIIT Pet & Oxford-Flower-102 \\
    \hline
    \textbf{OPSC+PA (ours)} & \textbf{80.65} & \textbf{84.26} & \textbf{85.75} & \textbf{93.09}\\
    OPSC (ours) & 79.74 & 83.34 & 83.46 & 92.33\\
    PA (our previous \cite{xiao2015application})& 65.41 & 68.32 & 75.42 & 88.75\\
    \hline
  \end{tabular}
\end{table*} 
\begin{table*}[!ht]
  \centering
  \caption{Performances of patch filtering. 
  }
  \label{fine-tune}
   \begin{tabular} {|p{2.6cm}<{\centering}|p{2.1cm}<{\centering}|p{2.1cm}<{\centering}|p{2.1cm}<{\centering}|p{2.5cm}<{\centering}|}
    \hline
    \multirow {2}{*}{Method} & \multicolumn{4}{c|}{Accuracy (\%)} \\
    \cline{2-5}
        & CUB-200-2011 & Cars-196 & Oxford-IIIT Pet &Oxford-Flower-102 \\ 
    \hline
    \textbf{ft-patches} & \textbf{80.82} & \textbf{86.79} & \textbf{88.14} & \textbf{94.70}\\
    ft-original & 80.11 & 85.76 & 87.52 & 93.84\\
    \hline
  \end{tabular}
\end{table*}
\item
We observe that ``Original+Part-level'' is better than ``Object-level+Part-level'', which shows the complementarity between ``Original'' and ``Part-level'' is stronger than that between ``Object-level'' and ``Part-level''. This is because: 1) Parts are selected based on the obtained object regions, which leads to the fact that selected parts are mostly inside object regions and cover the whole object regions. This causes that the complementarity between object and part is small. 2) Object localization may be wrong and cause that the localized object region does not contain the whole object, some areas of this object are outside the object region. These areas may be helpful for classification, which are not in the localized object region but in the original image. 3) Image also includes the information of background, which may be helpful for classification to a certain extent. So ``Original+Part-level'' can provide more supplementary information than ``Object-level+Part-level'', thus achieves better performance. Totally, ``Original+Object-level+Part-level'' further improves the classification accuracy due to the complementary information among image, object and part.

\item
Fig. \ref{partresult} shows some results of object localization and part selection by our OPAM approach. The first row denotes the original images, the second row denotes the localized objects of original images via object-level attention model, and the third and fourth rows denote the selected discriminative parts via part-level attention model. For CUB-200-2011, Cars-196 and Oxford-IIIT Pet datasets, the selected parts have explicit semantic meanings, where the third row denotes the head of object and the fourth denotes the body. For Oxford-Flower-102 dataset, there are two types of images: one contains only one flower, and the other contains multiple flowers. For the images containing only one flower, object means the flower and parts mean the discriminative regions of the flower, such as petal, flower bud or receptacle. For the images containing multiple flowers, object means the salient flower or the entirety of all flowers in image, and parts mean the discriminative regions of the flower or one single individual of the flowers. Our proposed approach is effective in both two cases, which localizes the discriminative objects and parts as well as learns fine-grained features to boost classification accuracy.
It is noted that neither object nor part annotations are used in our OPAM approach, which avoids the heavy labor consumption of labeling as well as pushes fine-grained image classification towards practical applications.
\end{itemize}

\subsubsection{Effectivenesses of object-part spatial constraint model and part alignment} 
Compared with our previous conference paper \cite{xiao2015application}, which only performs part alignment for selecting discriminative parts, we further employ object-part spatial constraint model to drive the discriminative part selection. 
The object spatial constraint ensures selected parts with high representativeness, while part spatial constraint eliminates redundancy and enhances discrimination of selected parts. Both of them are jointly employed to exploit the subtle and local discrimination for distinguishing the subcategories. 
In Fig. \ref{cluster} and Table \ref{SCPA}, ``OPSC'' refers to the object-part spatial constraint model, ``PA'' refers to part alignment which is adopted by our previous conference paper \cite{xiao2015application}, and ``OPSC+PA'' refers to combining the above two ones, which is adopted by our OPAM approach. 
From the left columns of four datasets in Fig. \ref{cluster}, we can see that only performing part alignment in part-level attention model without object-part spatial constraint causes the selected parts: 
(1) have large areas of background noise but small areas of object, (2) have large overlap with each other which leads to the redundant information. 
From Table \ref{SCPA}, we can see that the classification accuracies of parts selected by object-part spatial constraint model (``OPSC'') are better than parts selected with part alignment (``PA'') on all 4 datasets. Besides, applying part alignment on the basis of object-part spatial constraint further improves the classification performance. This verifies that aligning discriminative parts with the same semantic meaning together can further improve the results of part-level attention model.  

\subsubsection{Effectiveness of patch filtering} 
Through patch filtering in the object-level attention model, some image patches are selected from the candidate image patches. These patches are relevant to objects, and provide multiple views and scales of original images. These relevant patches are used to train \emph{ClassNet} to boost the effectiveness of \emph{ClassNet}. In Table \ref{fine-tune}, ``ft-patches'' refers to fine-tuning on image patches selected through patch filtering in object-level attention model and ``ft-original'' refers to fine-tuning only on original images. The results are the classification accuracies of prediction on original images. 
Fine-tuning on the selected image patches achieves better accuracy due to the effectiveness of multi-view and multi-scale feature learning based on the patch filtering in our OPAM approach.

\section{Conclusion}
In this paper, the OPAM approach has been proposed for weakly supervised fine-grained image classification, which jointly integrates two level attention models: object-level localizes objects of images, and part-level selects discriminative parts of objects. The two level attentions jointly improve the multi-view and multi-scale feature learning and enhance their mutual promotions. 
Besides, part selection is driven by the object-part spatial constraint model, which combines two spatial constraints: object spatial constraint ensures the high representativeness of selected parts, and part spatial constraint eliminates redundancy and enhances discrimination of selected parts. 
Combination of the two spatial constraints promotes the subtle and local discrimination localization.  
Importantly, our OPAM avoids the heavy labor consumption of labeling to march toward practical application. 
Comprehensive experimental results show the effectiveness of our OPAM approach compared with more than 10 state-of-the-art methods on 4 widely-used datasets.  

The future work lies in two aspects: First, we will focus on learning better fine-grained representation via more effective and precise part localization methods. Second, we will also attempt to apply semi-supervised learning into our work to make full use of large amounts of web data. Both of them will be employed to further improve the fine-grained image classification performance.



\ifCLASSOPTIONcaptionsoff
  \newpage
\fi

\balance 
{\small
\bibliographystyle{plain}
\bibliography{reference}

\begin{thebibliography}{10}

\bibitem{wah2011caltech}
Catherine Wah, Steve Branson, Peter Welinder, Pietro Perona, and Serge
  Belongie.
\newblock The caltech-ucsd birds-200-2011 dataset.
\newblock 2011.

\bibitem{krause20133d}
Jonathan Krause, Michael Stark, Jia Deng, and Li~Fei-Fei.
\newblock 3d object representations for fine-grained categorization.
\newblock {\em International Conference of Computer Vision Workshop (ICCV)},
  pages 554--561, 2013.

\bibitem{parkhi2012cats}
Omkar~M Parkhi, Andrea Vedaldi, Andrew Zisserman, and CV~Jawahar.
\newblock Cats and dogs.
\newblock {\em IEEE Conference on Computer Vision and Pattern Recognition
  (CVPR)}, pages 3498--3505, 2012.

\bibitem{Nilsback2008Automated}
Maria~Elena Nilsback and Andrew Zisserman.
\newblock Automated flower classification over a large number of classes.
\newblock {\em Sixth Indian Conference on Computer Vision, Graphics \& Image
  Processing}, pages 722--729, 2008.

\bibitem{maji2013fine}
Subhransu Maji, Esa Rahtu, Juho Kannala, Matthew Blaschko, and Andrea Vedaldi.
\newblock Fine-grained visual classification of aircraft.
\newblock {\em arxiv:1306.5151}, 2013.

\bibitem{zhang2014part}
Ning Zhang, Jeff Donahue, Ross Girshick, and Trevor Darrell.
\newblock Part-based r-cnns for fine-grained category detection.
\newblock {\em European conference on computer vision (ECCV)}, pages 834--849,
  2014.

\bibitem{zhang2016picking}
Xiaopeng Zhang, Hongkai Xiong, Wengang Zhou, Weiyao Lin, and Qi~Tian.
\newblock Picking deep filter responses for fine-grained image recognition.
\newblock {\em IEEE Conference on Computer Vision and Pattern Recognition
  (CVPR)}, pages 1134--1142, 2016.

\bibitem{zhang2016fused}
Xiaopeng Zhang, Hongkai Xiong, Wengang Zhou, and Qi~Tian.
\newblock Fused one-vs-all features with semantic alignments for fine-grained
  visual categorization.
\newblock {\em IEEE Transactions on Image Processing (TIP)}, 25(2):878--892,
  2016.

\bibitem{uijlings2013selective}
Jasper~RR Uijlings, Koen~EA van~de Sande, Theo Gevers, and Arnold~WM Smeulders.
\newblock Selective search for object recognition.
\newblock {\em International Journal of Computer Vision (IJCV)},
  104(2):154--171, 2013.

\bibitem{girshick2014rich}
Ross Girshick, Jeff Donahue, Trevor Darrell, and Jitendra Malik.
\newblock Rich feature hierarchies for accurate object detection and semantic
  segmentation.
\newblock {\em IEEE Conference on Computer Vision and Pattern Recognition
  (CVPR)}, pages 580--587, 2014.

\bibitem{branson2014bird}
Steve Branson, Grant Van~Horn, Serge Belongie, and Pietro Perona.
\newblock Bird species categorization using pose normalized deep convolutional
  nets.
\newblock {\em arxiv:1406.2952}, 2014.

\bibitem{krause2015fine}
Jonathan Krause, Hailin Jin, Jianchao Yang, and Li~Fei-Fei.
\newblock Fine-grained recognition without part annotations.
\newblock {\em IEEE Conference on Computer Vision and Pattern Recognition
  (CVPR)}, pages 5546--5555, 2015.

\bibitem{zhou2016fine}
Feng Zhou and Yuanqing Lin.
\newblock Fine-grained image classification by exploring bipartite-graph
  labels.
\newblock {\em IEEE Conference on Computer Vision and Pattern Recognition
  (CVPR)}, pages 1124--1133, 2016.

\bibitem{zhang2016weakly}
Yu~Zhang, Xiu-Shen Wei, Jianxin Wu, Jianfei Cai, Jiangbo Lu, Viet-Anh Nguyen,
  and Minh~N Do.
\newblock Weakly supervised fine-grained categorization with part-based image
  representation.
\newblock {\em IEEE Transactions on Image Processing (TIP)}, 25(4):1713--1725,
  2016.

\bibitem{xiao2015application}
Tianjun Xiao, Yichong Xu, Kuiyuan Yang, Jiaxing Zhang, Yuxin Peng, and Zheng
  Zhang.
\newblock The application of two-level attention models in deep convolutional
  neural network for fine-grained image classification.
\newblock {\em IEEE Conference on Computer Vision and Pattern Recognition
  (CVPR)}, pages 842--850, 2015.

\bibitem{lowe2004distinctive}
David~G Lowe.
\newblock Distinctive image features from scale-invariant keypoints.
\newblock {\em International Journal of Computer Vision (IJCV)}, 60(2):91--110,
  2004.

\bibitem{xie2014spatial}
Lingxi Xie, Qi~Tian, Meng Wang, and Bo~Zhang.
\newblock Spatial pooling of heterogeneous features for image classification.
\newblock {\em IEEE Transactions on Image Processing (TIP)}, 23(5):1994--2008,
  2014.

\bibitem{gao2014learning}
Shenghua Gao, Ivor Wai-Hung Tsang, and Yi~Ma.
\newblock Learning category-specific dictionary and shared dictionary for
  fine-grained image categorization.
\newblock {\em IEEE Transactions on Image Processing (TIP)}, 23(2):623--634,
  2014.

\bibitem{zhang2013deformable}
Ning Zhang, Ryan Farrell, Forrest Iandola, and Trevor Darrell.
\newblock Deformable part descriptors for fine-grained recognition and
  attribute prediction.
\newblock {\em International Conference of Computer Vision (ICCV)}, pages
  729--736, 2013.

\bibitem{simon2015neural}
Marcel Simon and Erik Rodner.
\newblock Neural activation constellations: Unsupervised part model discovery
  with convolutional networks.
\newblock {\em International Conference of Computer Vision (ICCV)}, pages
  1143--1151, 2015.

\bibitem{jaderberg2015spatial}
Max Jaderberg, Karen Simonyan, Andrew Zisserman, et~al.
\newblock Spatial transformer networks.
\newblock {\em Neural Information Processing Systems (NIPS)}, pages 2017--2025,
  2015.

\bibitem{huang2016part}
Shaoli Huang, Zhe Xu, Dacheng Tao, and Ya~Zhang.
\newblock Part-stacked cnn for fine-grained visual categorization.
\newblock {\em IEEE Conference on Computer Vision and Pattern Recognition
  (CVPR)}, pages 1173--1182, 2016.

\bibitem{zhang2016spda}
Han Zhang, Tao Xu, Mohamed Elhoseiny, Xiaolei Huang, Shaoting Zhang, Ahmed
  Elgammal, and Dimitris Metaxas.
\newblock Spda-cnn: Unifying semantic part detection and abstraction for
  fine-grained recognition.
\newblock {\em IEEE Conference on Computer Vision and Pattern Recognition
  (CVPR)}, pages 1143--1152, 2016.

\bibitem{wang2015multiple}
Dequan Wang, Zhiqiang Shen, Jie Shao, Wei Zhang, Xiangyang Xue, and Zheng
  Zhang.
\newblock Multiple granularity descriptors for fine-grained categorization.
\newblock {\em International Conference on Computer Vision (ICCV)}, pages
  2399--2406, 2015.

\bibitem{lin2015bilinear}
Tsung-Yu Lin, Aruni RoyChowdhury, and Subhransu Maji.
\newblock Bilinear cnn models for fine-grained visual recognition.
\newblock {\em International Conference of Computer Vision (ICCV)}, pages
  1449--1457, 2015.

\bibitem{zhaosurvey}
Bo~Zhao, Jiashi Feng, Xiao Wu, and Shuicheng Yan.
\newblock A survey on deep learning-based fine-grained object classification
  and semantic segmentation.
\newblock {\em International Journal of Automation and Computing}, pages 1--17,
  2017.

\bibitem{sermanet2014attention}
Pierre Sermanet, Andrea Frome, and Esteban Real.
\newblock Attention for fine-grained categorization.
\newblock {\em arxiv:1412.7054}, 2014.

\bibitem{zhou2016cvpr}
Bolei Zhou, Aditya Khosla, Agata Lapedriza, Aude Oliva, and Antonio Torralba.
\newblock Learning deep features for discriminative localization.
\newblock {\em IEEE Conference on Computer Vision and Pattern Recognition
  (CVPR)}, 2016.

\bibitem{liu2016fully}
Xiao Liu, Tian Xia, Jiang Wang, and Yuanqing Lin.
\newblock Fully convolutional attention localization networks: Efficient
  attention localization for fine-grained recognition.
\newblock {\em arxiv:1603.06765}, 2016.

\bibitem{xie2016interactive}
Lingxi Xie, Liang Zheng, Jingdong Wang, Alan~L Yuille, and Qi~Tian.
\newblock Interactive: Inter-layer activeness propagation.
\newblock {\em IEEE Conference on Computer Vision and Pattern Recognition
  (CVPR)}, pages 270--279, 2016.

\bibitem{zhao2016diversified}
Bo~Zhao, Xiao Wu, Jiashi Feng, Qiang Peng, and Shuicheng Yan.
\newblock Diversified visual attention networks for fine-grained object
  classification.
\newblock {\em arxiv:1606.08572}, 2016.

\bibitem{imagenet_cvpr09}
Jia Deng, Wei Dong, Richard Socher, Li-Jia Li, Kai Li, and Li~Fei-Fei.
\newblock {ImageNet: A Large-Scale Hierarchical Image Database}.
\newblock {\em IEEE Conference on Computer Vision and Pattern Recognition
  (CVPR)}, pages 248--255, 2009.

\bibitem{nadler2006diffusion}
Boaz Nadler, Stephane Lafon, Ioannis Kevrekidis, and Ronald~R Coifman.
\newblock Diffusion maps, spectral clustering and eigenfunctions of
  fokker-planck operators.
\newblock {\em Advances in Neural Information Processing Systems (NIPS)}, pages
  955--962, 2006.

\bibitem{simonyan2014very}
Karen Simonyan and Andrew Zisserman.
\newblock Very deep convolutional networks for large-scale image recognition.
\newblock {\em arxiv:1409.1556}, 2014.

\bibitem{ioffe2015batch}
Sergey Ioffe and Christian Szegedy.
\newblock Batch normalization: Accelerating deep network training by reducing
  internal covariate shift.
\newblock {\em International Conference on Machine Learning (ICML)}, pages
  448--456, 2015.

\bibitem{kohavi1995study}
Ron Kohavi et~al.
\newblock A study of cross-validation and bootstrap for accuracy estimation and
  model selection.
\newblock {\em International Joint Conference on Artificial Intelligence
  (IJCAI)}, 14(2):1137--1145, 1995.

\bibitem{xu2017friend}
Zhe Xu, Dacheng Tao, Shaoli Huang, and Ya~Zhang.
\newblock Friend or foe: Fine-grained categorization with weak supervision.
\newblock {\em IEEE Transactions on Image Processing (TIP)}, 26(1):135--146,
  2017.

\bibitem{zhang2016detecting}
Luming Zhang, Yang Yang, Meng Wang, Richang Hong, Liqiang Nie, and Xuelong Li.
\newblock Detecting densely distributed graph patterns for fine-grained image
  categorization.
\newblock {\em IEEE Transactions on Image Processing (TIP)}, 25(2):553--565,
  2016.

\bibitem{yao2016coarse}
Hantao Yao, Shiliang Zhang, Yongdong Zhang, Jintao Li, and Qi~Tian.
\newblock Coarse-to-fine description for fine-grained visual categorization.
\newblock {\em IEEE Transactions on Image Processing (TIP)}, 25(10):4858--4872,
  2016.

\bibitem{cui2015fine}
Yin Cui, Feng Zhou, Yuanqing Lin, and Serge Belongie.
\newblock Fine-grained categorization and dataset bootstrapping using deep
  metric learning with humans in the loop.
\newblock {\em arxiv:1512.05227}, 2015.

\bibitem{xu2016webly}
Zhe Xu, Shaoli Huang, Ya~Zhang, and Dacheng Tao.
\newblock Webly-supervised fine-grained visual categorization via deep domain
  adaptation.
\newblock {\em IEEE Transactions on Pattern Analysis and Machine Intelligence
  (TPAMI)}, 2016.

\bibitem{lin2015deep}
Di~Lin, Xiaoyong Shen, Cewu Lu, and Jiaya Jia.
\newblock Deep lac: Deep localization, alignment and classification for
  fine-grained recognition.
\newblock {\em IEEE Conference on Computer Vision and Pattern Recognition
  (CVPR)}, pages 1666--1674, 2015.

\bibitem{berg2013poof}
Thomas Berg and Peter Belhumeur.
\newblock Poof: Part-based one-vs.-one features for fine-grained
  categorization, face verification, and attribute estimation.
\newblock {\em IEEE Conference on Computer Vision and Pattern Recognition
  (CVPR)}, pages 955--962, 2013.

\bibitem{xie2013hierarchical}
Lingxi Xie, Qi~Tian, Richang Hong, Shuicheng Yan, and Bo~Zhang.
\newblock Hierarchical part matching for fine-grained visual categorization.
\newblock {\em International Conference of Computer Vision (ICCV)}, pages
  1641--1648, 2013.

\bibitem{xie2015hyper}
Saining Xie, Tianbao Yang, Xiaoyu Wang, and Yuanqing Lin.
\newblock Hyper-class augmented and regularized deep learning for fine-grained
  image classification.
\newblock {\em IEEE Conference on Computer Vision and Pattern Recognition
  (CVPR)}, pages 2645--2654, 2015.

\bibitem{krause2014learning}
Jonathan Krause, Timnit Gebru, Jia Deng, Li-Jia Li, and Li~Fei-Fei.
\newblock Learning features and parts for fine-grained recognition.
\newblock {\em International Conference on Pattern Recognition (ICPR)}, pages
  26--33, 2014.

\bibitem{wang2016mining}
Yaming Wang, Jonghyun Choi, Vlad Morariu, and Larry~S Davis.
\newblock Mining discriminative triplets of patches for fine-grained
  classification.
\newblock {\em IEEE Conference on Computer Vision and Pattern Recognition
  (CVPR)}, pages 1163--1172, 2016.

\bibitem{wang2016weakly}
Yaming Wang, Vlad~I Morariu, and Larry~S Davis.
\newblock Weakly-supervised discriminative patch learning via cnn for
  fine-grained recognition.
\newblock {\em arxiv:1611.09932}, 2016.

\bibitem{wang2010locality}
Jinjun Wang, Jianchao Yang, Kai Yu, Fengjun Lv, Thomas Huang, and Yihong Gong.
\newblock Locality-constrained linear coding for image classification.
\newblock {\em IEEE Conference on Computer Vision and Pattern Recognition
  (CVPR)}, pages 3360--3367, 2010.

\bibitem{xie2015image}
Lingxi Xie, Richang Hong, Bo~Zhang, and Qi~Tian.
\newblock Image classification and retrieval are one.
\newblock {\em ACM on International Conference on Multimedia Retrieval}, pages
  3--10, 2015.

\bibitem{azizpour2015generic}
Hossein Azizpour, Ali Sharif~Razavian, Josephine Sullivan, Atsuto Maki, and
  Stefan Carlsson.
\newblock From generic to specific deep representations for visual recognition.
\newblock {\em IEEE Conference on Computer Vision and Pattern Recognition
  Workshops (CVPR)}, pages 36--45, 2015.

\bibitem{qian2015fine}
Qi~Qian, Rong Jin, Shenghuo Zhu, and Yuanqing Lin.
\newblock Fine-grained visual categorization via multi-stage metric learning.
\newblock {\em IEEE Conference on Computer Vision and Pattern Recognition
  (CVPR)}, pages 3716--3724, 2015.

\bibitem{iscen2015comparison}
Ahmet Iscen, Giorgos Tolias, Philippe-Henri Gosselin, and Herv{\'e} J{\'e}gou.
\newblock A comparison of dense region detectors for image search and
  fine-grained classification.
\newblock {\em IEEE Transactions on Image Processing (TIP)}, 24(8):2369--2381,
  2015.

\bibitem{murray2014generalized}
Naila Murray and Florent Perronnin.
\newblock Generalized max pooling.
\newblock {\em IEEE Conference on Computer Vision and Pattern Recognition
  (CVPR)}, pages 2473--2480, 2014.

\bibitem{bo2013multipath}
Liefeng Bo, Xiaofeng Ren, and Dieter Fox.
\newblock Multipath sparse coding using hierarchical matching pursuit.
\newblock {\em IEEE Conference on Computer Vision and Pattern Recognition
  (CVPR)}, pages 660--667, 2013.

\bibitem{angelova2013efficient}
Anelia Angelova and Shenghuo Zhu.
\newblock Efficient object detection and segmentation for fine-grained
  recognition.
\newblock {\em IEEE Conference on Computer Vision and Pattern Recognition
  (CVPR)}, pages 811--818, 2013.

\bibitem{huang2017pbc}
Chao Huang, Hongliang Li, Yurui Xie, Qingbo Wu, and Bing Luo.
\newblock Pbc: Polygon-based classifier for fine-grained categorization.
\newblock {\em IEEE Transactions on Multimedia (TMM)}, 19(4):673--684, 2017.

\bibitem{xie2017towards}
Lingxi Xie, Jingdong Wang, Weiyao Lin, Bo~Zhang, and Qi~Tian.
\newblock Towards reversal-invariant image representation.
\newblock {\em International Journal of Computer Vision (IJCV)},
  123(2):226--250, 2017.

\bibitem{sharif2014cnn}
Ali Sharif~Razavian, Hossein Azizpour, Josephine Sullivan, and Stefan Carlsson.
\newblock Cnn features off-the-shelf: an astounding baseline for recognition.
\newblock {\em IEEE Conference on Computer Vision and Pattern Recognition
  Workshops (CVPR)}, pages 806--813, 2014.

\bibitem{plant}
Anelia Angelova and Shenghuo Zhu.
\newblock Efficient object detection and segmentation for fine-grained
  recognition.
\newblock {\em IEEE Conference on Computer Vision and Pattern Recognition
  (CVPR)}, pages 811--818, 2013.

\bibitem{azizpour2012object}
Hossein Azizpour and Ivan Laptev.
\newblock Object detection using strongly-supervised deformable part models.
\newblock {\em European conference on computer vision (ECCV)}, pages 836--849,
  2012.

\bibitem{szegedy2015going}
Christian Szegedy, Wei Liu, Yangqing Jia, Pierre Sermanet, Scott Reed, Dragomir
  Anguelov, Dumitru Erhan, Vincent Vanhoucke, and Andrew Rabinovich.
\newblock Going deeper with convolutions.
\newblock {\em IEEE Conference on Computer Vision and Pattern Recognition
  (CVPR)}, pages 1--9, 2015.

\bibitem{chatfield2014return}
Ken Chatfield, Karen Simonyan, Andrea Vedaldi, and Andrew Zisserman.
\newblock Return of the devil in the details: Delving deep into convolutional
  nets.
\newblock {\em arxiv:1405.3531}, 2014.

\end{thebibliography}
}

%








\end{document}